\documentclass[a4paper,fleqn]{cas-dc}

\usepackage[T1]{fontenc}
\usepackage{placeins}
\makeatletter
\AtEndDocument{\clearpage
  \immediate\write\@auxout{\string\gdef\string\lastpage{\number\numexpr\value{page}-1}}}
\makeatother
\usepackage[authoryear,longnamesfirst]{natbib}
\usepackage{amsmath}
\usepackage{booktabs}
\usepackage{multirow}
\usepackage{graphicx}
\graphicspath{{figs/}}

\begin{document}
\let\WriteBookmarks\relax
\def\floatpagepagefraction{1}
\def\textpagefraction{.001}


\shortauthors{P. Parajuli and S. Dahal}

\title[mode=title]{Fine-resolution land-use segmentation and synthetic
image--label augmentation in the Kathmandu Valley}

\author[1]{Pranil Parajuli}[type=editor,
                            auid=000,
                            bioid=1,
                            orcid=0009-0008-7044-0585]
\fnmark[1]
\ead{pparajuli15@gmail.com}
\credit{Conceptualization, Methodology, Software, Validation, Formal analysis,
Investigation, Data curation, Visualization, Writing -- original draft,
Writing -- review and editing}

\affiliation[1]{organization={Department of Electronics and Computer
                Engineering, Institute of Engineering, Thapathali Campus,
                Tribhuvan University},
                city={Kathmandu},
                postcode={44600},
                state={Bagmati Province},
                country={Nepal}}

\author[1]{Sugam Dahal}[orcid=0009-0004-0786-6656]
\cormark[1]
\fnmark[1]
\ead{dahalsugam24@gmail.com}
\credit{Conceptualization, Methodology, Software, Validation, Formal analysis,
Investigation, Data curation, Visualization, Writing -- original draft,
Writing -- review and editing}

\author[1]{Kiran Chandra Dahal}
\ead{dahalkc@tcioe.edu.np}
\credit{Supervision}

\cortext[1]{Corresponding author.}
\fntext[1]{Pranil Parajuli and Sugam Dahal contributed equally to this work and
share first authorship. The work was carried out jointly throughout, without a
division of tasks between them.}
\begin{abstract}
Parcel-scale land-use data is missing for most rapidly urbanizing Global South cities,  as current segmentation architectures rely on multispectral imagery and planned layouts. We introduce a hand-annotated six-class land-use corpus covering 202 km²  of Kathmandu Valley across 408 fine-resolution RGB scenes. The dataset is uniquely challenging: RGB spectral similarity is high, class imbalance approaches 100:1, and key structural features like rivers measure just two to three pixels wide at a standard hierarchical decoder’s prediction stride. Evaluating five architectures reveals that standard random data splitting conceals massive spatial leakage. Switching to a strict sequence-block split drops performance by up to 0.095 mIoU. This penalty exceeds the performance gap between the architectures themselves. We modified the SegFormer baseline by integrating a half-resolution residual detail path, boundary-reliability reweighting, a topology-aware loss, and a presence-balanced sampler. Adding only 1.9\% more parameters, the enhanced architecture improves mean IoU across every test fold, with gains concentrated in rare, thin-structure classes. However, both models achieve identical results on the external LoveDA benchmark. Finally, augmenting with diffusion-generated pairs at matched compute degraded accuracy across all architectures, especially harming rare classes it aimed to improve. This demonstrates that blindly incorporating synthetic data can severely damage performance, highlighting a critical need to rigorously evaluate and filter generated samples before training.
\end{abstract}


\maketitle

\section{Introduction}\label{sec:introduction}

The Kathmandu Valley is among the fastest urbanising metropolitan regions in South Asia. Four decades of satellite observation have documented the conversion of
country's agricultural core to built-up land \citep{thapa2011urbangrowth,
ishtiaque2017kathmandu}. The downstream impacts are no longer projections. The benefits of nature are declining across the valley floor
\citep{rimal2022ecosystem}.Notably, the heat level depends on the layout and composition of land cover, not just the total built-up volume.
\citep{karunaratne2022suhi}. Effective spatial planning requires detailed maps of land-use layout rather than aggregate estimates of built-up area. This data deficit exceptionally impacts effective urban design. Global satellite maps routinely offer 10-meter resolution \citep{brown2022dynamicworld, zanaga2021worldcover},
a single 10-meter pixel in Kathmandu covers both a house and its adjacent street, obscuring vital details.
The gap is structural. The morphology of rapidly urbanising cities in the Global South is dominated by small, irregular, densely packed parcels, which is why a separate literature on mapping unplanned settlement from fine-resolution imagery exists at all \citep{kuffer2016slums, wurm2019slums}. Nepal sits almost entirely outside its empirical base.

Semantic segmentation is the mature technical response \citep{long2015fcn, xie2021segformer, guo2022segnext}, and because a diffusion model can synthesise
the \emph{label} as readily as the image \citep{toker2024satsynth}, annotation
scarcity now looks attackable generatively. But the benchmarks that validate segmentation encode a particular geography: German towns
\citep{rottensteiner2012isprs}, global satellite tiles
\citep{demir2018deepglobe} and Chinese cities \citep{wang2021loveda}, all with planned grids, regular footprints and usually a near-infrared band. Transferring those segmentations to the Kathmandu Valley crosses a wider gap. Three properties of our data, all measured in Section~\ref{sec:data}, obstruct it: every class pair is spectrally inseparable in RGB, the median road is 2.1 pixels wide at a hierarchical decoder's prediction stride, and the imbalance ratio is 94:1 with river absent from four fifths of the tiles. Flawed reporting protocols create an additional hurdle. Overhead imagery is spatially autocorrelated. A random validation set contains neighbours of training tiles and inflates measured accuracy \citep{kattenborn2022spatially, ploton2020spatial}, and random splits are yet the norm in this domain.

This study evaluates these constraints using a purpose built dataset, specifically testing generative augmentation to mitigate the third challenge. We present four main contributions:

\begin{enumerate}[(i)]
\item A six-class land-use dataset covering 202.4~km$^2$ of the Kathmandu Valley, released with annotations, fold definitions and metric records. We also document the dataset's main challenges, including the loss of export geometry and the subsequent recovery of the acquisition scale using photogrammetric measurements (Section~\ref{sec:data}).
\item We compare five architectures under one protocol across two partitioning regimes, covering thirty fold-runs with results reported per fold. We also quantify how much the partitioning choice alone affects reported performance (Sections~\ref{sec:setup}--\ref{sec:results}).
\item We develop a transformer designed for thin structures and coarse supervision that outperforms its backbone in all six fold-runs with only 1.9~\% more parameters. A leave-one-out ablation shows that one component has no measurable effect, while the four individual contributions explain less than a third of the combined improvement (Sections~\ref{sec:method} and~\ref{sec:discussion}).
\item We test diffusion generated image--label pairs as training data under a pre-specified, compute matched setup, with target classes and fold selected before evaluating the results. The synthetic data provides no improvement and instead harms rare thin classes, for which we identify the underlying cause (Sections~\ref{sec:results} and~\ref{sec:discussion}).
\end{enumerate}

\section{Related work}\label{sec:related}

\subsection{Segmentation of fine-resolution overhead imagery}
\label{sec:related-seg}

Dense labelling became practical with fully convolutional networks \citep{long2015fcn} and the encoder--decoder architecture \citep{ronneberger2015unet}, later advancing toward hierarchical attention \citep{liu2021swin, xie2021segformer} and large-kernel convolutional attention \citep{guo2022segnext} to address the loss of boundary detail when downsampling for larger receptive fields. In overhead imagery, materials rather than objects define the semantics \citep{yuan2021review}, leading to remote-sensing architectures designed around this distinction \citep{wang2022unetformer, li2021abcnet}. These architectures assume multispectral input, a planned grid, or both, while their efficient hierarchical decoders predict at reduced resolution. This works well for objects much wider than the output stride but poorly for smaller structures.

\subsection{Thin structures and imbalance}\label{sec:related-thin}

Roads and rivers are where that trade is worst \citep{zhou2018dlinknet, dang2021msresnet}. Pixel-wise objectives are unsuitable for these structures because a fragmented road can score well at the pixel level despite losing its network connectivity. The clDice metric addresses this by computing Dice between predicted and reference skeletons \citep{shit2021cldice}. Two assumptions do not hold here. First, these methods are designed for \emph{binary} extraction on curated data, while our rare classes account for just 0.60~\% and 1.57~\% of pixels in a six-class problem. Second, they assume accurate boundaries, but our hand-drawn polygons are least reliable in the very regions where topology-aware losses concentrate their gradients. Class imbalance is commonly addressed through reweighting \citep{cui2019classbalanced} or overlap-based optimisation \citep{berman2018lovasz}, but their interactions remain unexplored. Section~\ref{sec:disc-ablation} shows that the four individual effects sum to less than half their combined effect.

\subsection{Evaluation under spatial dependence}\label{sec:related-eval}

\citet{kattenborn2022spatially} showed that spatial autocorrelation between training and validation samples inflates the reported performance of convolutional networks on remote-sensing data, while \citet{karasiak2022spatial} demonstrated similar problems in accuracy assessment. The ecological literature had reached the same conclusion earlier and more strongly \citep{ploton2020spatial}. The counter argument is equally established. \citet{wadoux2021spatialcv} argue that spatial cross-validation is unsuitable for assessing map accuracy and may be overly pessimistic under probability sampling. Meanwhile, \citet{meyer2021aoa} approach the problem through the model's area of applicability. Following established guidance on accuracy assessment \citep{olofsson2014goodpractices}, we treat the two protocols as bounds on the true performance. Reporting only one would be misleading, so we present both.

\subsection{Generative synthesis of image--label pairs}
\label{sec:related-gen}

Diffusion models learn to invert a fixed noising process \citep{ho2020ddpm}, made practical by deterministic sampling \citep{song2021ddim}, better parameterisation \citep{salimans2022progressive, lin2024commondiffusion} and a compressed latent space \citep{rombach2022ldm}. ControlNet advances spatial conditioning by copying a frozen backbone's encoder into a trainable branch connected through zero initialised convolutions \citep{zhang2023controlnet}. Under small data constraints, this approach is combined with low-rank adaptation \citep{hu2022lora}. A six-class land-use palette is not a control signal the pretrained backbone understands. Remote sensing quickly adopted diffusion models \citep{liu2024dmrs, khanna2024diffusionsat, tang2024crsdiff}, but these studies used $10^{5}$--$10^{7}$ images compared with our 804 and generated \emph{imagery} from externally supplied layouts. A generator trained on real layouts cannot create new ones, which is precisely what a corpus of 408 scenes needs. An alternative is to generate images and labels together using diffusion over discrete data \citep{chen2023analogbits}, with SatSynth serving as the closest predecessor to our work \citep{toker2024satsynth}.

Whether synthetic data helps remains debated, with improvements mostly seen in low data settings and sometimes reversed by distribution mismatch \citep{he2023syntheticready}. Evaluating its quality is also difficult. Fr\'echet and kernel Inception distances \citep{heusel2017fid, binkowski2018kid} measure appearance rather than label accuracy and are strongly influenced by ImageNet class priors that do not apply to nadir imagery \citep{kynkaanniemi2023imagenetclasses}. With small datasets, generators may also reproduce their training samples \citep{somepalli2023forgery}. Our evaluation addresses these three issues rather than focusing only on appearance.

\section{Study area and dataset}\label{sec:data}

\subsection{Study area and imagery}\label{sec:data-area}
\begin{figure*}[!t]
  \centering
  \includegraphics[width=\linewidth]{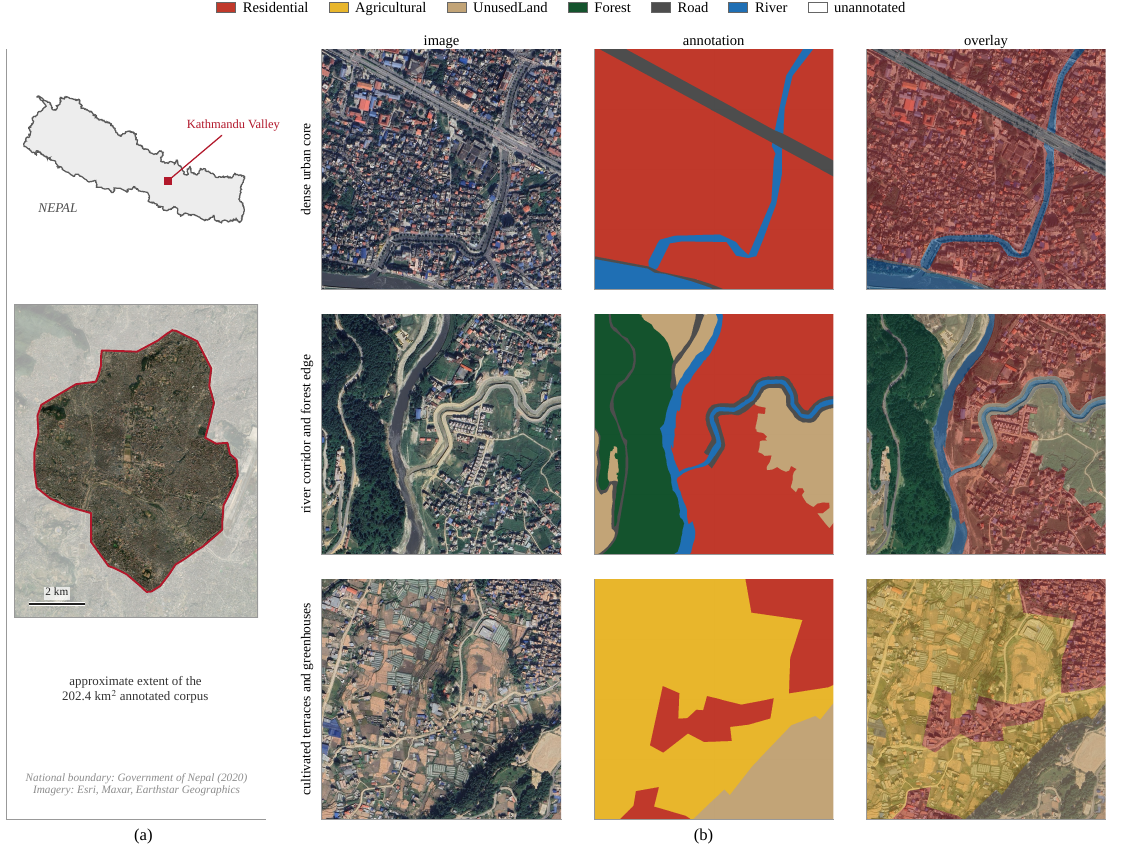}
  \caption{The corpus. (a)~Approximate extent of the 202.4~km$^2$
covered by 408 annotated tiles. Tile footprints were not preserved,
so a georeferenced tile index is unavailable.  (b)~Three
representative tiles from the dense urban core, river corridor and
cultivated fringe, shown as images, annotations and overlays.}
  \label{fig:studyarea}
\end{figure*}

The study area covers the urban and peri-urban regions of the Kathmandu Valley in Bagmati Province, Nepal. It includes dense historic settlements, low-rise residential areas, river corridors, remaining agricultural terraces and forested hills, often within the same scene. The dataset covers 202.4~km$^2$ across 408 contiguous, non-overlapping tiles (Fig.~\ref{fig:studyarea}). Imagery is the Google
Satellite basemap exported through a QGIS print-layout atlas as
2345~$\times$~2351 pixel 8-bit RGB rasters. We do not specify the sensor or orbital parameters because the basemap is a mosaicked, tone mapped product with unknown per-tile provenance. Near-infrared data is also unavailable.

The atlas project was lost after annotation, along with the tile footprints, layout scale and export resolution. Although the imagery cannot recover position, it allows us to estimate the scale. One tile shows a stadium with a football pitch whose markings have fixed regulatory dimensions regardless of pitch size. The penalty area measures exactly 40.32~$\times$~16.50~m, while the goal area measures 18.32~$\times$~5.50~m. We locate the pitch automatically and rotate it upright using its fitted angle. The painted lines are then isolated by subtracting a 21-pixel local median, with their positions measured from sub-pixel profile peaks. The four estimates range from 0.2989--0.3087~m per pixel and give a baseline-length-weighted scale of $0.3016 \pm 0.0049$~m per pixel ($\pm$1.6~\%). Section~\ref{sec:data-scale} details the error budget and three independent checks, including one from another tile. These checks were not used in the original calculation and confirm that the scale remains consistent across the export. We use a native ground sampling distance of 0.300~m per pixel, giving each tile a footprint of 703.5~$\times$~705.3~m. At the 512~$\times$~512 raster used by the models, the working resolution is 1.374~m per pixel. All measurements reported in metres carry a $\pm$1.6~\% uncertainty, which is smaller than the corresponding variation between tiles.

\subsection{Recovery of the acquisition scale}\label{sec:data-scale}

\begin{figure}[!t]
  \centering
    \includegraphics[width=\linewidth]{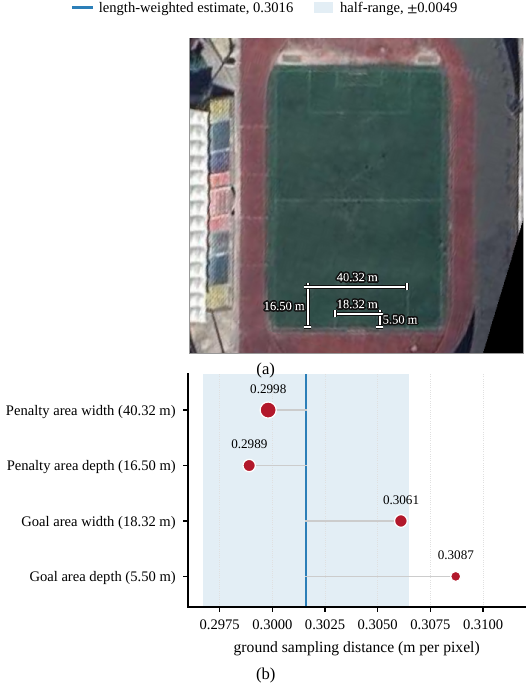}
  \caption{Recovering the acquisition scale. (a)~Football pitch rotated upright with four
regulation fixed controls marked. (b)~Ground sampling distance estimated
from these controls with marker size reflecting baseline length.}
  \label{fig:scale}
\end{figure}

Table~\ref{tbl:a1} and Fig.~\ref{fig:scale} show the four reference distances, each fixed by the Laws of the Game regardless of the pitch's overall dimensions. We combine them using length based weights, giving the total known distance in metres divided by the total measured pixels. This is the least-squares scale estimate when localisation error is constant for each line. The residuals show the pattern expected from a constant per line error. The two longer baselines give closely matching estimates, while the shorter pair shows greater variation. All four controls remain within 0.89~pixels of the combined estimate. This agrees with the expected accuracy of a sub-pixel intensity centroid estimator for painted lines about 12~cm wide. We therefore consider $\pm$1.6~\% a realistic uncertainty bound rather than an overly cautious estimate. Three independent checks, none used in the original derivation, also pass. Halfway line symmetry agrees within 0.73 pixels over a 344.84-pixel baseline, and the estimated pitch dimensions of 104.0~$\times$~67.7~m fall within regulation limits. A marked long-course swimming pool in a \emph{different} tile gives a result within $+$4.1~\%. A standard 1:3000 sheet scanned at 254~dpi (0.1~mm per pixel) would give a nominal resolution of 0.300~m per pixel. This differs from our estimate by just 0.53~\%, providing strong supporting evidence, though not definitive proof.

\begin{table}[t]
\caption{Ground sampling distance estimated from four fixed dimension
controls. Error shows the deviation in pixels along each control's baseline.}
\label{tbl:a1}
\begingroup
\setlength{\tabcolsep}{3pt}
\begin{tabular*}{\tblwidth}{@{}L@{\extracolsep{\fill}}CCCC@{}}
\toprule
Control & True (m) & Measured (px) & GSD & Error (px) \\
\midrule
Penalty area width & 40.32 & 134.50 & 0.2998 & $-$0.82 \\
Penalty area depth & 16.50 &  55.20 & 0.2989 & $-$0.49 \\
Goal area width    & 18.32 &  59.85 & 0.3061 & $+$0.89 \\
Goal area depth    &  5.50 &  17.81 & 0.3087 & $+$0.42 \\
\midrule
Length-weighted    & --- & --- & 0.3016 & --- \\
Half-range         & --- & --- & $\pm$0.0049 & --- \\
\bottomrule
\end{tabular*}
\endgroup
\end{table}

\subsection{Annotation and class schema}\label{sec:data-annot}

We annotated every tile in CVAT using polygon shapes at roughly
10--15 minutes per tile, exported in COCO format and rasterised to a fixed
six-colour palette. There was no second review pass and inter-annotator
agreement was not measured. Three operational definitions are important for understanding the results. \emph{Road} includes both carriageways and footpaths because they cannot be reliably distinguished at this resolution. The annotated roads therefore form corridors with a median width of 11.5~m. \emph{Forest} includes only large, continuous tree patches, excluding urban and street trees. This distinction explains a confusion discussed in Section~\ref{sec:disc-interp}. \emph{Unused land} includes areas showing no evidence of previous or planned cultivation, along with vacant plots between urban buildings. As a result, unused land appears alongside residential areas in 86~\% of tiles containing it.

We examined all 2{,}249{,}342{,}760 label pixels and found exactly seven colours: six classes and pure black, with no near-palette values. The black pixels represent annotation errors rather than a separate class. They account for 0.347~\% of all label pixels across 333 of 408 tiles, with one tile containing 49.19~\% unannotated pixels. We exclude these pixels from all loss calculations and evaluation metrics. Downsampling labels to 512 pixels using nearest-neighbour interpolation changes no class proportion by more than 0.07~\% relative and removes no class from any tile. Class balance can therefore be reported at the working resolution.

\subsection{What makes the corpus hard}\label{sec:data-stats}

Table~\ref{tbl:classstats} shows the class distribution, with an imbalance ratio of 94.1:1 and an effective class count, $\exp(H)$, of 3.42 out of 6. However, class presence across tiles matters more than overall pixel share. River appears in only 83 of 408 tiles and covers at most 9.26~\% of any tile. Residential areas also appear in 96.4~\% of river containing tiles, meaning river pixels must compete with a dominant class during loss calculation.

\begin{table}[t]
\caption{Per-class statistics across 408 scenes at 512-pixel resolution.
  ``Worst JM'' indicates the lowest Jeffries--Matusita separability
  from another class on a 0--2 scale. ``px @$H/4$'' gives
  the median width at the hierarchical decoder's prediction stride.}
\label{tbl:classstats}
\begin{tabular*}{\tblwidth}{@{}LRRRRCC@{}}
\toprule
Class & Share & Area & Tiles & Worst & Width & px @ \\
      & (\%)  & (km$^2$) &   & JM    & (m)   & $H/4$ \\
\midrule
Residential  & 56.02 & 113.41 & 360 & 0.204 & --- & --- \\
Agricultural & 18.10 &  36.64 & 191 & 0.051 & --- & --- \\
Unused land  & 12.97 &  26.25 & 225 & 0.051 & --- & --- \\
Forest       & 10.39 &  21.02 & 132 & 0.476 & --- & --- \\
Road         &  1.57 &   3.18 & 214 & 0.290 & 11.54 & 2.10 \\
River        &  0.60 &   1.20 &  83 & 0.290 & 14.82 & 2.70 \\
\midrule
Unannotated  &  0.35 &   0.74 & 333 & --- & --- & --- \\
\bottomrule
\end{tabular*}
\end{table}

\begin{figure*}[!t]
  \centering
  \includegraphics[width=\linewidth]{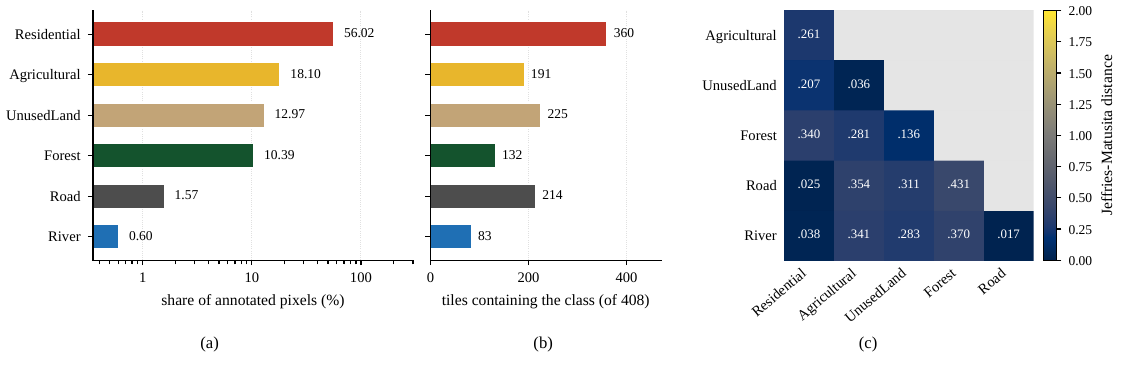}
  \caption{Properties of the corpus that make it hard. (a)~Share of annotated pixels by
  class, logarithmic axis. (b)~Number of tiles containing each class, total 408.
  (c)~Jeffries--Matusita separability for every class pair, from per-class
  Gaussians in RGB.}
  \label{fig:difficulty}
\end{figure*}

Two measurements help explain the results that follow (Fig.~\ref{fig:difficulty}). We represent each class as a multivariate Gaussian in RGB and calculate the Jeffries--Matusita distances. All distances are below 1.33, with only road and forest exceeding 1.0 at 1.322. Unused land and agricultural land are especially similar, with a distance of just 0.051. Their RGB signatures are nearly identical, leaving models to rely on texture and surrounding context. Skeletonising the class masks gives a median road width of 8.4 pixels across 198.15~km of centreline and a median river width of 10.8 pixels across 56.28~km. At quarter resolution prediction stride, these widths fall to 2.10 and 2.70 pixels, with the thinnest 10~\% of roads below one pixel.

\subsection{Partitioning protocol}\label{sec:data-split}

Each scene exists three times on disk (original, horizontal mirror and vertical
mirror), giving 1224 image--mask pairs from 408 independent scenes. Pixel equality tests confirmed that the mirrored images are exact copies. We therefore calculate all statistics using the 408 original scenes and keep each scene's three variants within the same split. Flip augmentation is also disabled during training. Section~\ref{sec:results-leakage} examines what happens when this restriction is removed.

\begin{figure}[!t]
  \centering
  \includegraphics[width=\linewidth]{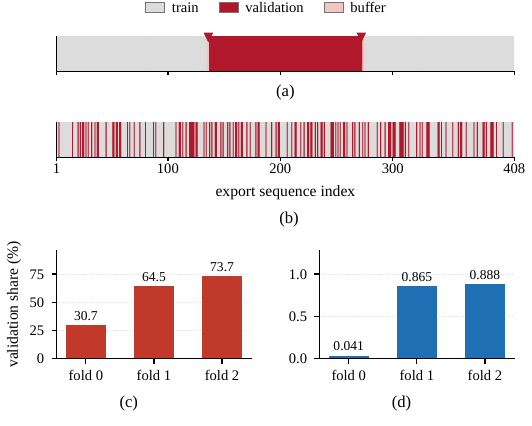}
  \caption{Two partitioning regimes for the 408-scene export sequence.
  (a)~Sequence-block fold~1 showing dropped two-tile buffers.
  (b)~Random split of the same scenes. (c,~d)~Residential and river
  proportions in each block fold's validation set.}
  \label{fig:partitions}
\end{figure}

Since the tile coordinates cannot be recovered, true geo-blocked cross-validation is not possible. Instead, we use consecutive runs of the export sequence as a spatial proxy, excluding a two-tile buffer from the training side of each block boundary. We refer to this approach as \emph{sequence-block cross-validation (export-order proxy)} throughout, rather than spatial cross-validation. We also use a random split of the same scenes to examine how much performance improves with less restrictive partitioning. Each mode has three folds, containing 268--272 training scenes and 136 validation scenes (Fig.~\ref{fig:partitions}).

We test the reliability of this proxy rather than simply assuming it works. All classes show positive lag-one autocorrelation in per-tile class share along the export axis ranging from 0.411 to 0.790. Consecutive tiles are also twice as similar in composition as random pairs ($L_1$ distance 0.5531 versus 1.0630). Perceptual hashing of all 83{,}028 scene pairs finds no near-duplicates below a Hamming distance of 8/64, with a minimum distance of 14. This rules out leakage caused by visually duplicated scenes. Two limitations remain. Adjacency perpendicular to the scan axis is uncontrolled, and some tiles were manually deleted and renumbered before annotation. The proxy therefore \emph{under-}estimates true spatial leakage making our reported gap a lower bound. The three block folds also cover different parts of the valley which matters when interpreting their results. Validation residential shares are 30.66~\%, 64.49~\% and 73.65~\%, while river shares are 0.041~\%, 0.865~\% and 0.888~\% across 3, 38 and 42 tiles. Fold~0 cannot reliably measure river performance. These block folds represent three distinct problems rather than repeated samples of the same one.

\section{Methodology}\label{sec:method}

\begin{figure*}[!t]
  \centering
  \includegraphics[width=\linewidth]{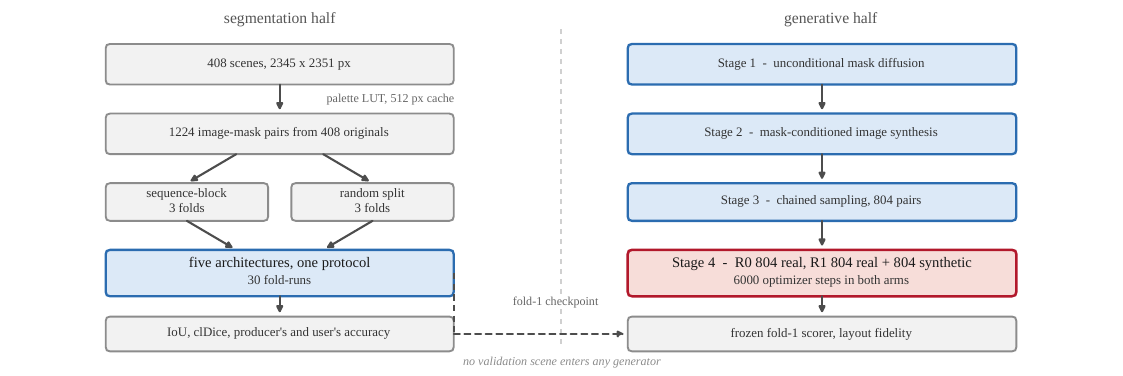}
 \caption{The study pipeline. Left: segmentation benchmark comparing five
  architectures under one protocol across two partitioning regimes,
  with thirty fold-runs. Right: generative pipeline covering unconditional
  mask diffusion, mask-conditioned image synthesis and compute-matched
  downstream evaluation.}
  \label{fig:pipeline}
\end{figure*}

The study has two parts following the same protocol (Fig.~\ref{fig:pipeline}). First, we train five segmentation architectures, including our modified compact transformer under both partitioning regimes. Next, we train a two-stage diffusion generator using the training scenes from one fold. Its synthetic image--label pairs are then added to the training data for all five architectures with computational budgets kept equal.

\subsection{The proposed segmentation model}\label{sec:method-tc}

The proposed model adds three components to SegFormer-B0 backbone
\citep{xie2021segformer}, each independently switchable so that each has a
clean ablation, together adding 0.078~M parameters or 1.9~\% of the backbone.

\emph{A residual half-resolution detail path.} The backbone combines features from its four encoder stages at $H/4$ and produces logits at that resolution. As Table~\ref{tbl:classstats} shows, this is at or below the resolution needed to represent both linear classes. To recover finer details, a shallow three-layer convolutional branch processes the raw image at stride 2. Its output is concatenated with the projected decoder features at $H/2$, fused, and converted into a six-channel residual added to the upsampled backbone logits. The final convolution uses an initial standard deviation of $10^{-4}$, while a learned scalar gate starts at 0.1. At step zero, the model matches the vanilla backbone to four decimal places. Disabling the branch reproduces the backbone exactly, providing an exact control for the corresponding ablation.

\begin{figure*}[!t]
  \centering
  \includegraphics[width=\linewidth]{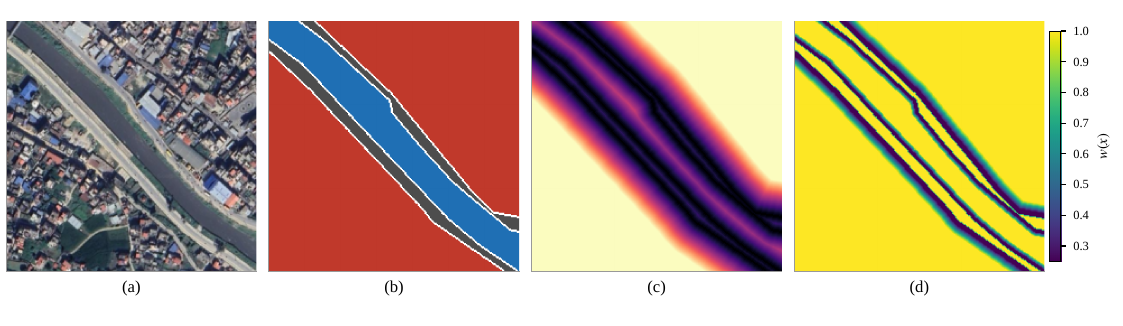}
  \caption{Per-pixel reliability weights from Eq.~\eqref{eq:csa} for a crop
  containing a road and large residential block. (a)~Image.
  (b)~Annotation with boundary set $B$ marked in white.
  (c)~Distance transform of the complement
  of $B$. 
  (d)~Resulting weight field, floored at 0.25.}
  \label{fig:weight}
\end{figure*}

\emph{Coarse-supervision-aware reweighting.} The polygons were drawn quickly without review. Pixels just inside and outside their boundaries often look nearly identical. We therefore assign each pixel a reliability weight based on annotation geometry, without learned parameters. Let $B$ contain pixels whose four neighbourhood has a different label, excluding image borders. Define $D(x)$ as the Euclidean distance transform of the complement of $B$. For each connected component of each class, the local half-width is $R = 2\,\mathrm{mean}(D)$. We then calculate
\begin{equation}\label{eq:csa}
  w(x) = \mathrm{clip}\!\left(\frac{D(x)}{b_c}, w_{\min}, 1\right),
  \quad
  b_c = \mathrm{clip}\!\left(\tau_c R,\, 1\ \mathrm{px},\, d_{\max}\right),
\end{equation}
Here, $\tau_c = 0.35$ for the two linear classes and $0.60$ for the four area classes, with $w_{\min} = 0.25$ and a metric cap $d_{\max}$ of 14.07 ground metres. The key difference is that the unreliable band depends on \emph{local object width} rather than a fixed pixel distance. Every road pixel is within roughly 4.2 pixels of a boundary. A fixed width band would therefore push road and river weights towards $w_{\min}$, weakening the very classes we aim to improve. The metric cap prevents this imbalance, remaining inactive for linear classes while limiting the area classes (Fig.~\ref{fig:weight}).

\emph{Topology-aware supervision and presence-balanced sampling.}
We train the model using a combination of weighted cross-entropy,
Lov\'asz-softmax, and soft centreline Dice losses:
\begin{equation}\label{eq:loss}
  \mathcal{L} = \mathcal{L}^{w}_{\mathrm{CE}}
  + \lambda(e)\,\mathcal{L}_{\mathrm{Lov}}
  + \mu(e)\,\mathcal{L}_{\mathrm{clDice}}^{\{\mathrm{road},\,\mathrm{river}\}},
\end{equation}
For cross-entropy, we apply effective number class weights
\citep{cui2019classbalanced}, label smoothing of 0.05,
and the pixel weights defined in Eq.~\eqref{eq:csa}.
We add Lov\'asz-softmax \citep{berman2018lovasz}
as a surrogate for the Jaccard index, increasing
its coefficient $\lambda$ gradually to 0.75 during
the first ten epochs. Soft centreline Dice loss
\citep{shit2021cldice} applies only to roads and rivers.
It is enabled at epoch 20, when $\mu$ increases
from 0 to 0.30. Earlier in training, predicted regions
may be too sparse for skeleton based supervision
to offer meaningful guidance.

\begin{figure*}[!t]
  \centering
  \includegraphics[width=\linewidth]{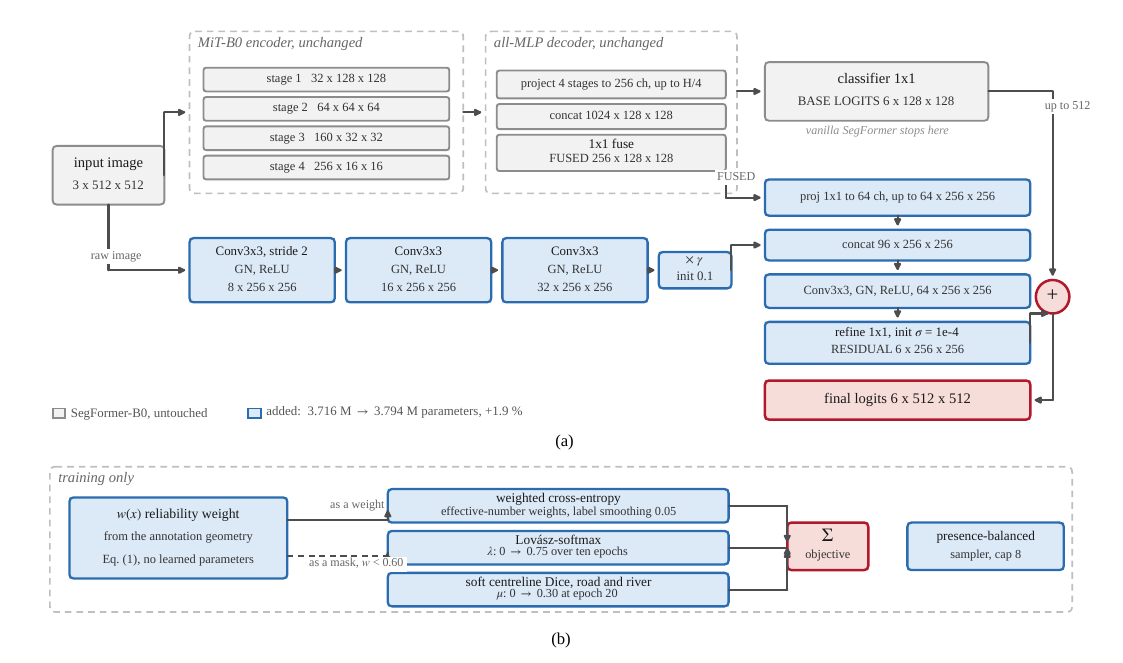}
  \caption{The proposed model. (a)~Forward pass with the unchanged SegFormer-B0
  backbone (grey) and three added components (blue). The detail path
  adds a gated six-channel residual to the upsampled backbone logits.
  (b)~Training objective using reliability weights in cross-entropy
  and a hard mask in the Lov\'asz-softmax term.}
  \label{fig:model}
\end{figure*}

We apply reliability weights differently to cross-entropy and Lov\'asz-softmax. Cross-entropy operates at the pixel level, allowing each pixel's contribution to be weighted independently. Lov\'asz-softmax, however, ranks prediction errors making direct pixel-wise weighting unsuitable. Instead, we exclude pixels with reliability scores below 0.60 from this loss. Sampling also matters because present mode Lov\'asz-softmax only considers classes appearing in each batch. To address this, we assign each training scene a sampling probability proportional to the inverse tile presence frequency of its rarest class. We cap the resulting sampling weight at 8 to avoid repeatedly selecting scenes containing rare classes. Fig.~\ref{fig:model} shows the complete model.

\subsection{Synthetic image--label pair generation}\label{sec:method-gen}

\begin{figure*}[!t]
  \centering
  \includegraphics[width=\linewidth]{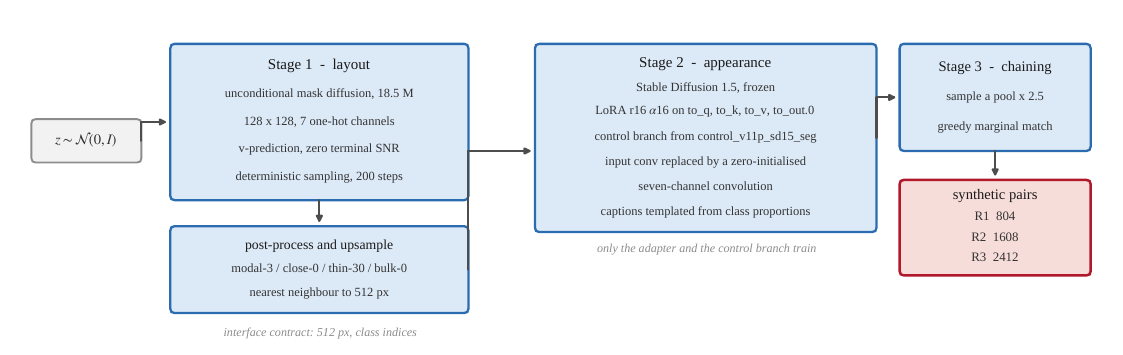}
  \caption{The two-stage generator. Stage~1 generates six-class label maps
  at 128~$\times$~128 through diffusion. Stage~2 uses a frozen Stable Diffusion~1.5
  backbone with low-rank adaptation and a control branch conditioned on
  class channels instead of a colour palette. Stage~3 chains the two
  and filters to the real class pixel marginal.}
  \label{fig:generator}
\end{figure*}

The generator is trained only on block fold~1's training scenes as the fold has enough river in its validation block for evaluation. Its segmentation checkpoint also serves as the frozen scorer. No validation scenes enter any stage, and matching hashes of the training scene list are checked throughout. Fig.~\ref{fig:generator} shows the three stages.

\begin{figure}[!t]
  \centering
  \includegraphics[width=\linewidth]{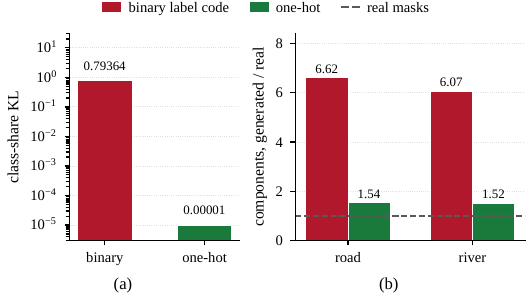}
  \caption{The label-representation ablation for the mask generator.
  (a)~Kullback--Leibler divergence between generated and real class share
  marginals on a logarithmic axis. (b)~Connected components per thin class
  relative to real masks, with the dashed line marking parity.}
  \label{fig:labelrep}
\end{figure}

\begin{figure}[!t]
  \centering
  \includegraphics[width=\linewidth]{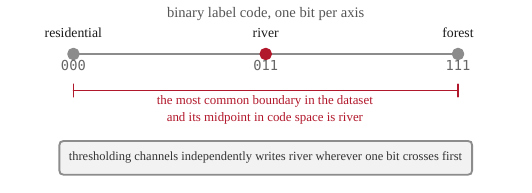}
  \caption{Why binary label encoding fails for thin classes. In the
  three channel code, river lies midway between residential
  and forest. This is the corpus's most common class boundary
  where a single bit transition incorrectly produces river labels.}
  \label{fig:labelcode}
\end{figure}

\emph{Stage 1, layout.} We train an unconditional denoising diffusion model \citep{ho2020ddpm} on six-class label maps, using 12~M parameters at 128~$\times$~128 resolution. Sampling uses 200 deterministic steps \citep{song2021ddim}, followed by nearest neighbour upsampling to 512 pixels. Labels are represented as seven one-hot channels, with $v$-prediction \citep{salimans2022progressive} and zero terminal signal-to-noise ratio \citep{lin2024commondiffusion}. We also report a failed alternative rather than omitting it. A three channel binary label code following \citet{toker2024satsynth} trained successfully but produced unusable masks. In this code, river lies exactly midway between residential and forest, the dataset's most common boundary. Independent channel thresholding therefore creates river pixels whenever one bit crosses its threshold first. All overproduced classes correspond to code midpoints, while underproduced classes supply their boundaries. This reveals a limitation of binary label coding when class boundaries are unevenly distributed (Figs.~\ref{fig:labelrep} and~\ref{fig:labelcode}).

\emph{Stage 2, appearance.} We use a frozen Stable Diffusion 1.5 backbone \citep{rombach2022ldm}, with rank-16 low-rank adaptation applied to its attention projections \citep{hu2022lora}. The control branch \citep{zhang2023controlnet} starts from a pretrained semantic ControlNet replacing its input convolution with a zero-initialised seven-channel convolution. This allows each class to enter as a separate channel. Unlike the standard pipeline, we avoid representing semantic conditions through colours. The pretrained ControlNet expects a specific colour palette, so changing it would introduce colours unrelated to its learned representations. Colour encoding can also create artificial relationships between classes. For instance, roads might appear closer to forests simply because their assigned colours are similar. Separate class channels avoid these misleading relationships. Captions follow short templates based on the class proportions in each mask.

\emph{Stage 3, chaining.} We combine the frozen models to generate $(\text{image}, \text{mask})$ pairs at one, two and three times the real training pool size. Each set uses independently sampled masks and image noise. We then greedily select samples from a candidate pool 2.5 times larger, matching the class pixel distribution of the real training data.

\subsection{The downstream experiment}\label{sec:method-stage4}

We evaluate each architecture under two training conditions. Arm~R0 trains on 804 real pairs from fold~1, while arm~R1 uses those same pairs plus 804 synthetic pairs. Both arms are validated on 408 real images from 136 held-out scenes, with no synthetic images included in validation.

We control training duration by \emph{optimizer steps} rather than epochs. With a batch size of 16, each pass through R0 takes 50 steps while R1 requires 100. Matching the number of epochs would therefore give R1 twice as many parameter updates, making it unclear whether improvements came from synthetic data or additional training. To avoid this, both arms run for 6000 steps, with the cosine schedule also ending at 6000. This gives R0 120 passes through the real data pool matching the benchmark training schedule, and R1 60 passes through its larger pool. Both process 96{,}000 sampled tiles. We keep all other training settings identical between the two arms. Class weights and sampling frequencies are calculated only from real training pairs ensuring that adding synthetic samples does not alter the loss configuration. Both arms follow the same augmentation policy, random seed, data split, warmup schedule, patience, and model initialisation. We also use identical bootstrap resample indices for the R0 and R1 comparison. 
We trained arm~R0 again under the new experimental setup instead of reusing the benchmark results. Across all five architectures, mIoU changed by no more than 0.006 compared with the original fold-1 results, leaving the conclusions from Table~\ref{tbl:main} unchanged.


\section{Experimental setup}\label{sec:setup}

We evaluated all five architectures under the same experimental protocol, changing only the architecture and learning rate between runs. Each model received 512$\times$512 RGB crops normalised using ImageNet statistics. The task involved six semantic classes and an ignore index. Training used AdamW with an effective batch size of 16 and weight decay of 0.01, excluding biases and normalisation layers. The encoder learning rate was one-tenth of the decoder learning rate.

The learning rate schedule began with 500 warmup steps followed by cosine decay to zero over 120 epochs. We stopped training early when validation mIoU showed no improvement of at least $\delta = 0.002$ for 25 epochs. We maintained exponential moving average weights with a decay of 0.999 and used them for all model evaluations. Label smoothing of 0.05 was applied only to cross-entropy. Training used fp16 mixed precision, while loss calculations remained in fp32.

During training, we augmented images with random rotations in 90-degree increments and brightness and contrast adjustments of $\pm$10\%. These augmentations were applied only during training, with no test-time augmentation at inference. All experiments ran on a single NVIDIA T4 GPU with 16~GB of memory and a 12 hour session limit.

We used a fixed random seed of 42 and a deterministic data pipeline. However, cuDNN kernel selection remained non-deterministic, so some variation between runs was still possible. All results come from a single seed without repeated independent runs. We therefore cannot fully assess the effects of random initialisation and training stochasticity on model performance. This is the main statistical limitation of our study.

We selected the four baselines to compare architectural differences rather than leaderboard rankings. Their parameter counts range from 3.7 to 12.5~M, keeping model capacities relatively similar. U-Net with a ResNet-18 encoder \citep{ronneberger2015unet,he2016resnet} serves as the conventional encoder--decoder baseline. UNetFormer \citep{wang2022unetformer} retains the same encoder but uses global--local transformer blocks in its decoder. This lets us examine the decoder's contribution without changing the encoder. We included SegNeXt-T \citep{guo2022segnext} to represent convolutional attention. SegFormer-B0 \citep{xie2021segformer} uses a hierarchical transformer and serves as the direct backbone comparison for our proposed model.

For UNetFormer, we used our own implementation rather than the original repository, which carries a GPL-3.0 licence. We also set the decoder learning rate to $3\times10^{-4}$ instead of the $6\times10^{-4}$ reported in the paper. Our UNetFormer results therefore reflect the implementation and training settings used in this study. They should not be interpreted as a direct reproduction of the original authors' findings.

We calculate all pixel-level metrics from one confusion matrix per run, excluding ignored pixels. The metrics include per-class and mIoU, overall accuracy, frequency-weighted IoU, Cohen's $\kappa$, and producer's and user's accuracy for each class. We report producer's and user's accuracy rather than macro F1 because F1 follows directly from IoU. Reporting recall and precision separately also helps distinguish between underprediction and overprediction of a class. For roads and rivers, we additionally measure centreline Dice \citep{shit2021cldice} using true skeletons. We exclude images where the class is absent from both the prediction and reference.

We measure layout fidelity by passing each synthetic image through a \emph{frozen} segmentation model and comparing its prediction against the mask used for generation. The resulting IoU is reported as a ratio to the model's performance ceiling on real images. We repeat the test with a second, independently trained segmentation model to check whether the findings remain consistent. For image realism, we primarily use kernel Inception distance \citep{binkowski2018kid}. We also report Fr\'echet Inception distance \citep{heusel2017fid}, although its estimates can be heavily biased with datasets of this size. To examine memorisation, we calculate nearest neighbour retrieval distances across feature, pixel, and mask spaces. We normalise each distance against its equivalent from held out \emph{real} tiles, allowing us to compare synthetic samples with real examples.

We estimate uncertainty in downstream performance differences using a paired bootstrap with 2000 resamples of the 136 validation scenes. Both experimental arms use the same resamples to preserve their pairing. The resulting intervals capture variation across validation scenes while keeping the training pool, random seed, and fold unchanged. They do not account for variation between training seeds. Since all five architectures are evaluated under identical fixed experimental conditions, their uncertainty intervals are also not independent.

\section{Results}\label{sec:results}

\subsection{Segmentation across the two partitioning regimes}
\label{sec:results-main}

Table~\ref{tbl:main} presents all thirty fold-runs. Across all five architectures, fold~1 consistently performs best, followed by fold~2 and fold~0. This pattern reflects differences in the data rather than the models.

\begin{table*}[t]
\caption{Per-fold mIoU for five architectures under both partitioning
regimes. Block-split SD measures variation across three regions,
not noise. Gap is random minus block.}
\label{tbl:main}
\begingroup
\setlength{\tabcolsep}{3pt}
\begin{tabular*}{\tblwidth}{@{}L@{\extracolsep{\fill}}CCCCCCCCCC@{}}
\toprule
\multirow{2}{*}{Architecture} & \multicolumn{4}{c}{Sequence-block split} &
\multicolumn{4}{c}{Random split} & \multirow{2}{*}{Gap} &
\multirow{2}{*}{SD ratio} \\
\cmidrule(lr){2-5}\cmidrule(lr){6-9}
 & Fold 0 & Fold 1 & Fold 2 & Mean (SD) & Fold 0 & Fold 1 & Fold 2 & Mean (SD) & & \\
\midrule
U-Net (R18)      & 0.3213 & 0.4516 & 0.4228 & 0.3986 (0.0684) & 0.4533 & 0.4465 & 0.4484 & 0.4494 (0.0035) & $+$0.0508 & 19.5$\times$ \\
UNetFormer (R18) & 0.2795 & 0.5069 & 0.4736 & 0.4200 (0.1228) & 0.5263 & 0.5007 & 0.4905 & 0.5058 (0.0184) & $+$0.0858 & 6.7$\times$ \\
SegFormer-B0     & 0.2877 & 0.4657 & 0.4351 & 0.3962 (0.0952) & 0.4948 & 0.4736 & 0.4611 & 0.4765 (0.0170) & $+$0.0803 & 5.6$\times$ \\
SegNeXt-T        & 0.3345 & \textbf{0.5701} & \textbf{0.5122} & \textbf{0.4723} (0.1228) & \textbf{0.5646} & 0.5498 & 0.5450 & 0.5531 (0.0102) & $+$0.0809 & 12.0$\times$ \\
Proposed         & \textbf{0.3403} & 0.5582 & 0.5061 & 0.4682 (0.1138) & 0.5525 & \textbf{0.5650} & \textbf{0.5717} & \textbf{0.5631} (0.0097) & $+$0.0949 & 11.7$\times$ \\
\bottomrule
\end{tabular*}
\endgroup
\end{table*}

\begin{figure*}[!t]
  \centering
  \includegraphics[width=\linewidth]{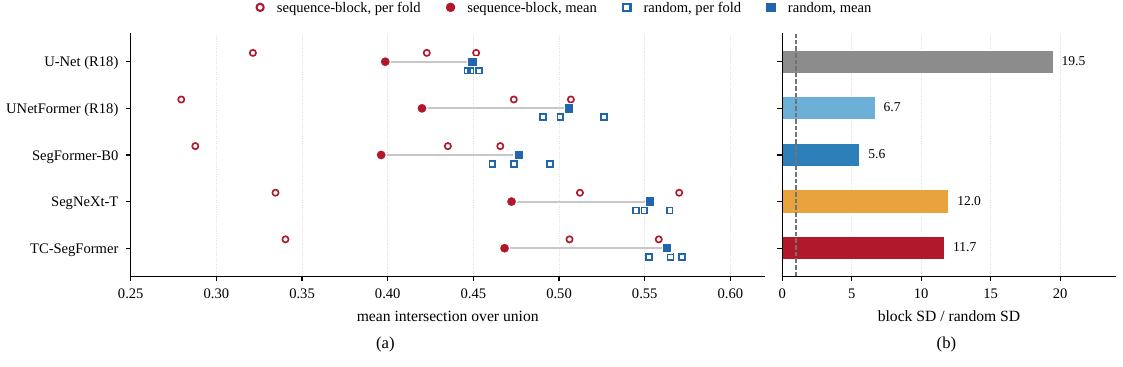}
  \caption{Effect of the partitioning protocol. (a)~mIoU for both regimes,
  with open markers showing three folds, filled markers showing
  regime means and connecting lines showing the gaps.
  (b)~Ratio of between-fold standard deviations under block
  and random splits.}
  \label{fig:gap}
\end{figure*}

The two right hand columns highlight the main methodological findings (Fig.~\ref{fig:gap}). Switching from random to sequence-block partitioning reduces mIoU by 0.0508 to 0.0949 across all architectures. This means five to nine points of apparent performance depend solely on whether spatially adjacent scenes can appear in both training and validation sets. For each architecture, the block-split standard deviation is 5.6 to 19.5 times greater than the random-split value. Both regimes use the same 408 scenes, models, seeds and training schedules, so nearly all variation between block folds reflects regional differences. These estimates are lower bounds because the export order proxy underestimates true spatial leakage.

\begin{figure}[!t]
  \centering
  \includegraphics[width=\linewidth]{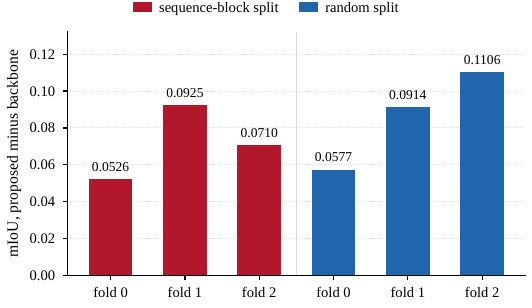}
  \caption{Performance difference between the proposed model and its
  SegFormer-B0 backbone across six fold-runs. Both share the same
  encoder, initialisation and training schedule, making this the only
  architectural comparison unaffected by capacity or pretraining differences.}
  \label{fig:vsbackbone}
\end{figure}

Only one pairwise comparison here is clean. The proposed model and
SegFormer-B0 share an encoder, an initialisation and a schedule; every other
pair confounds architecture with capacity, pretraining corpus and optimiser. On
that comparison the proposed model wins all six fold-runs, by $+$0.0526,
$+$0.0925 and $+$0.0710 under blocking, $+$0.0577, $+$0.0914 and $+$0.1106 at
random, or 11.7~\% to 24.0~\% in relative terms, for 0.078~M added parameters
(Fig.~\ref{fig:vsbackbone}). 
The remaining architectures cannot be ranked reliably. The top two differ by just 0.0041 in block mean and reverse positions under random partitioning, a difference too small to resolve with a single seed.

An external benchmark helps put these results into perspective. Using the same training harness, we obtain 0.4730 and 0.4718 mIoU on the public LoveDA benchmark (Section~\ref{sec:results-loveda}). Our corpus achieves 0.3962--0.4723 with sequence-block partitioning and 0.4494--0.5631 with random splitting. Under matched code and computational budgets, its difficulty is therefore comparable to an established seven-class benchmark. It is neither unusually easy nor exceptionally difficult. Section~\ref{sec:results-leakage} provides another, much larger comparison by measuring the effect of splitting flip-augmented copies of the same scene at the file level.

\subsection{Per-class results}\label{sec:results-class}

\begin{figure*}[!t]
  \centering
  \includegraphics[width=\linewidth]{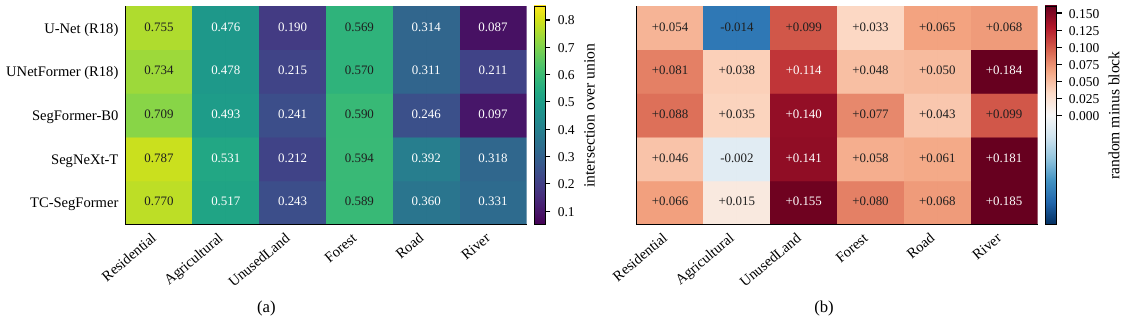}
  \caption{Per-class results by architecture. (a)~IoU averaged across
  the three sequence-block folds. (b)~Spatial penalty for the same
  cells, computed as random minus block.}
  \label{fig:perclass}
\end{figure*}

\begin{figure*}[p]
  \centering
  \includegraphics[width=\linewidth,height=0.90\textheight,keepaspectratio]%
    {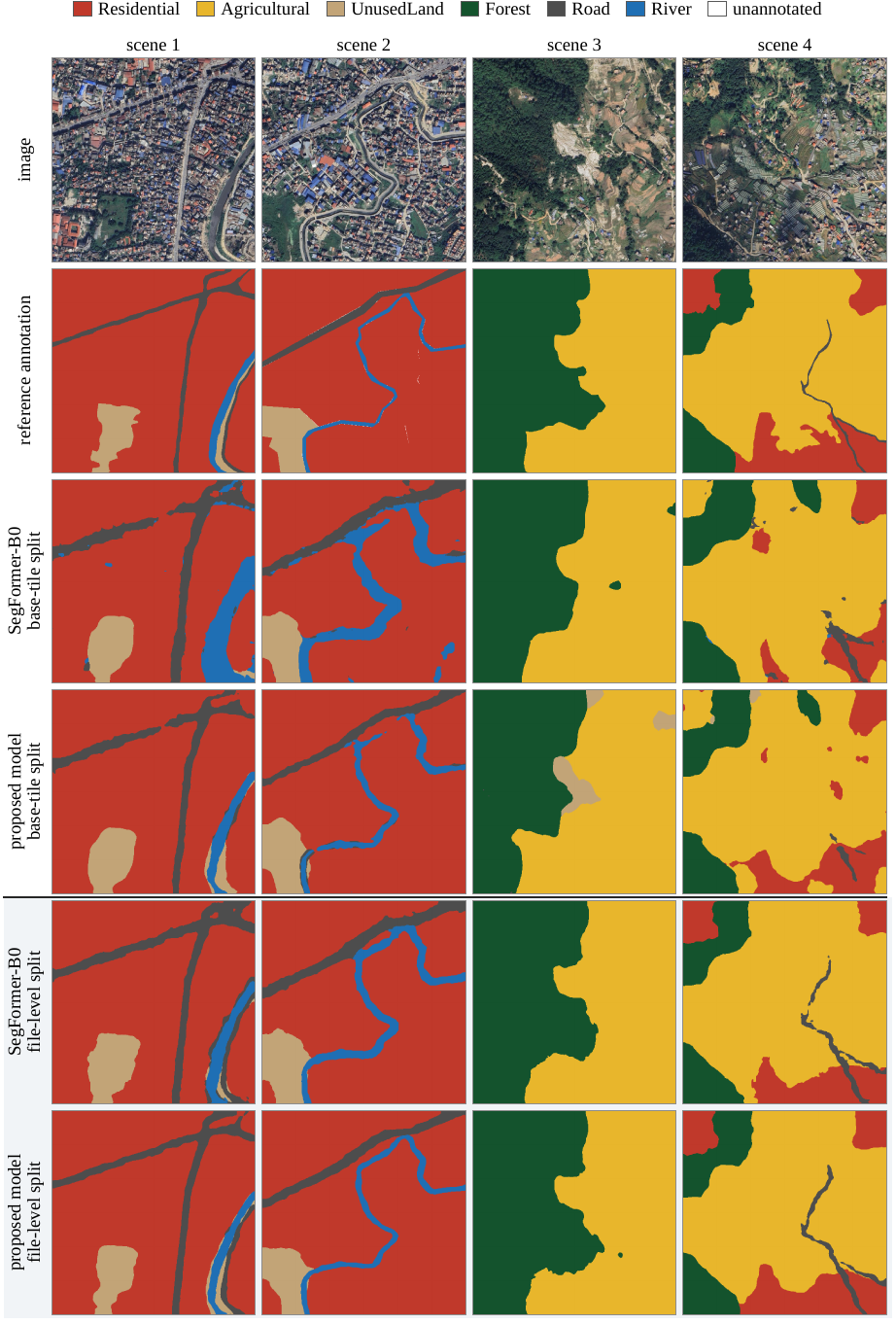}
  \caption{Qualitative comparison across four held-out scenes from block fold~1.
  Columns show the four scenes, with class colours indicated in the legend.
  Rows show the input images, reference annotations, predictions from
  SegFormer-B0 and the proposed model under the base-tile split, and the final two rows compare both models under the file-level split
  described in Section~\ref{sec:results-leakage}.}
\label{fig:qualitative}
\end{figure*}

The class-wise results in Table~\ref{tbl:perclass} reveal different patterns for the three weak classes. River scores vary considerably across architectures. Even with the block split, the highest mIoU is 3.8 times the lowest, while blocking lowers river IoU by 0.143. For agricultural land, the reduction is just 0.014. Road scores vary less across models and are less sensitive to blocking. This suggests that road errors may stem more from the difficulty of capturing narrow structures than from differences between spatial blocks. Unused land presents another challenge. Across all tested architectures, its IoU remains between 0.1899 and 0.2428 (Fig.~\ref{fig:perclass}). The predictions in Fig.~\ref{fig:qualitative} illustrate these errors. Unused land near the valley fringe is often mistaken for agricultural land, trees within built-up areas are labelled as forest, and river masks become fragmented along narrow sections.

\begin{table}[t]
\caption{Per-class IoU averaged over the three folds in each regime,
shown as the range across five architectures. ``Penalty'' is the
mean random minus block across the five.}
\label{tbl:perclass}
\begingroup
\setlength{\tabcolsep}{3pt}
\begin{tabular*}{\tblwidth}{@{}L@{\extracolsep{\fill}}CCC@{}}
\toprule
Class & Block & Random & Penalty \\
\midrule
Residential  & 0.7093--0.7872 & 0.7969--0.8354 & $+$0.0669 \\
Agricultural & 0.4757--0.5311 & 0.4614--0.5324 & $+$0.0144 \\
Forest       & 0.5694--0.5935 & 0.6027--0.6690 & $+$0.0593 \\
Road         & 0.2461--0.3918 & 0.2886--0.4526 & $+$0.0573 \\
Unused land  & 0.1899--0.2428 & 0.2893--0.3978 & $+$0.1301 \\
River        & 0.0868--0.3305 & 0.1546--0.5155 & $+$0.1433 \\
\bottomrule
\end{tabular*}
\endgroup
\end{table}

\subsection{External benchmark}\label{sec:results-loveda}

To evaluate the proposed model beyond our dataset, we tested it alongside its backbone on LoveDA \citep{wang2021loveda}. Both used the official training and validation split, the same harness, a fixed 20{,}000-step budget, and sliding window evaluation across all 1669 validation tiles. Published LoveDA results come from test-server evaluations after 80{,}000 iterations on datacentre hardware, so direct comparison is not appropriate. The meaningful comparison is between our two models which perform almost identically. SegFormer-B0 achieves a mIoU of 0.4730, compared with 0.4718 for the proposed model. Overall accuracy and $\kappa$ also differ by no more than 0.0034.

\begin{figure}[!t]
  \centering
  \includegraphics[width=\linewidth]{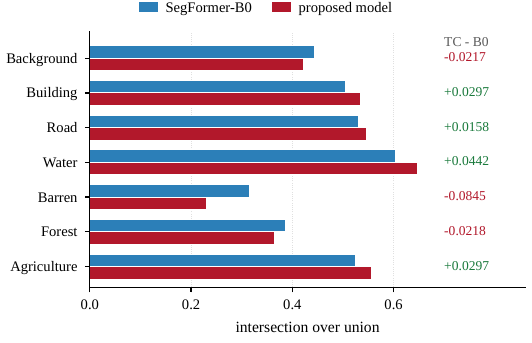}
  \caption{Per-class IoU on the LoveDA validation data after 20{,}000
  training steps. Mean performance is similar, but the
  proposed model improves water, building, agriculture and road
  while performing worse on barren, forest and background.}
  \label{fig:loveda}
\end{figure}

The similar overall scores hide differences across classes. The proposed model performs better on water ($+$0.0442), building ($+$0.0297), agriculture ($+$0.0297), and road ($+$0.0158). Water and road are also the two classes where its topology term is applied. However, performance drops on barren ($-$0.0845), forest, and background (both $-$0.022). Barren is structurally similar to unused land in LoveDA, and this architecture struggles most with this class on both datasets (Fig.~\ref{fig:loveda}).

The proposed model selected its checkpoint at step 19{,}000, compared with 9{,}000 for the backbone. This leaves two possibilities: training stopped while performance was still improving, or the model simply needed twice as many steps to reach similar performance. The available evidence cannot distinguish between them.

\subsection{Quality of the synthetic pairs}\label{sec:results-gen}

\begin{figure}[!t]
  \centering
  \includegraphics[width=\linewidth]{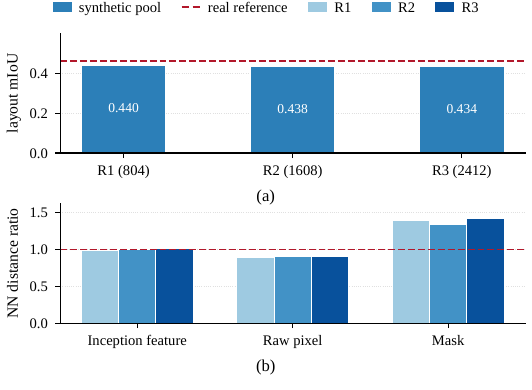}
  \caption{Synthetic pool quality at three set sizes.
  (a)~Layout fidelity, with the dashed line marking the frozen scorer's
  0.4646 ceiling on real images.
  (b)~Nearest-neighbour distance to training samples, relative to
  held-out real tiles (dashed).}
  \label{fig:poolquality}
\end{figure}

The generator produces image--label pairs with consistent labels rather than
copies of training samples. Layout fidelity reaches 0.4395, 0.4375 and
0.4337 mIoU for the one, two and three fold pools 
respectively. These correspond to 0.946, 0.942 and 0.933 of the frozen
scorer's 0.4646 ceiling on real imagery. A second, independently trained
scorer confirms this with pixel agreement of 0.890--0.904, while kernel
Inception distance ranges from 0.047--0.052. After distribution matching,
the class pixel marginal closely matches real training data with a
Kullback--Leibler divergence of $3\times10^{-5}$.
Nearest neighbour retrieval distances are 0.99--1.01 of the held-out-real
baseline in feature space and 1.34--1.42 in mask space. Only raw pixel
space falls below one, at 0.89--0.90. This reflects the generator learning
the dataset's colour statistics rather than reproducing training tiles
(Fig.~\ref{fig:poolquality}).

\begin{figure}[!t]
  \centering
  \includegraphics[width=\linewidth]{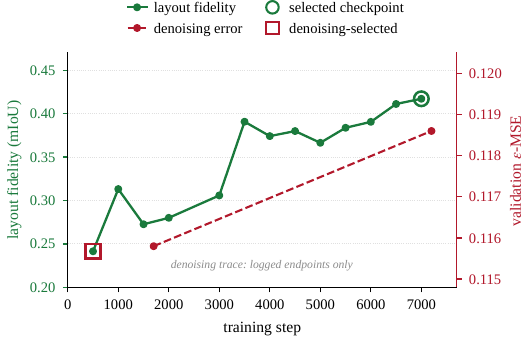}
 \caption{Denoising error selects the wrong image-stage checkpoint.
  Validation error increases even as layout fidelity improves
  from 0.24 to 0.42.}
  \label{fig:selector}
\end{figure}

Two findings concern the methodology rather than the numbers. First,
\emph{denoising loss is a poor criterion for selecting a conditional
generator's checkpoint}. During image-stage training, validation denoising
error increased steadily from 0.1158 at step 1700 to 0.1186 at step 7200,
even as layout fidelity improved from 0.24 to 0.42. An earlier run that
used denoising error for checkpoint selection achieved a fidelity ratio
of just 0.651 (Fig.~\ref{fig:selector}).

\begin{figure*}[!t]
  \centering
  \includegraphics[width=\linewidth]{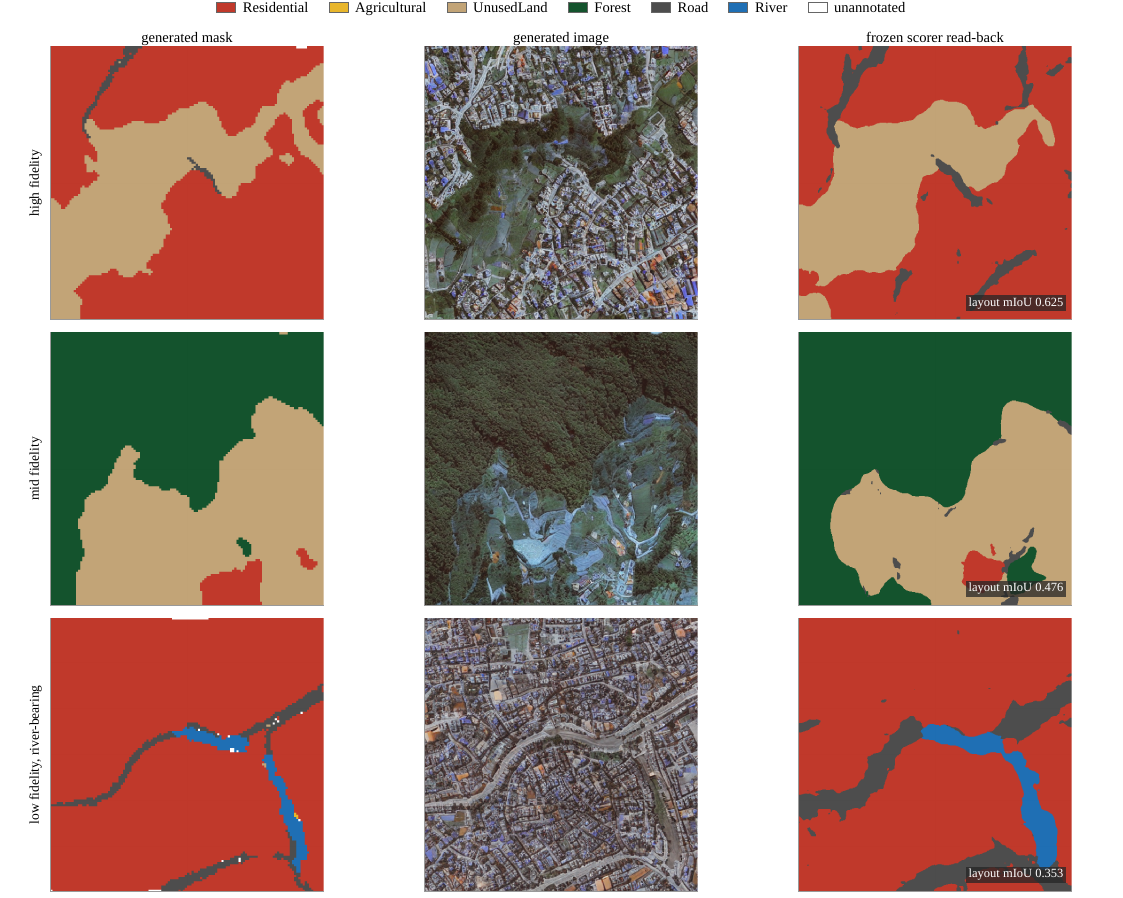}
  \caption{Three synthetic pairs with different fidelity levels.
  Each shows a generated mask, its corresponding synthetic image
  and the frozen scorer's prediction against the original mask.
  The rows show the error causing the decline in downstream
performance.}
  \label{fig:synthetic}
\end{figure*}

The layout stage independently showed the same pattern. Second, using the
same image model to paint \emph{real} validation masks produces a layout mIoU of 0.5238, compared with 0.4395 for generated masks.
This is 19~\% higher and exceeds the real-image ceiling. The remaining error
therefore comes mainly from layout generation, not image synthesis. However,
exceeding the ceiling is a warning, not necessarily an improvement. Generated
images are \emph{easier} to segment because they are smoother and more
prototypical than real images. This creates a distribution mismatch that
downstream experiments must account for. River also appears in 30--34~\% of
generated tiles compared with 16.8~\% in real data at the same pixel budget.
Augmentation could therefore increase rare class coverage rather than
repeatedly reproducing similar scenes. Fig.~\ref{fig:synthetic} shows pairs
with different levels of fidelity.

\subsection{Does synthetic augmentation help?}\label{sec:results-stage4}

The results show otherwise (Table~\ref{tbl:stage4}). Adding 804 synthetic
pairs at matched compute reduces mIoU for every
architecture with an average decline of $-$0.0140. Bootstrap intervals
exclude zero for two of the five losses. The same pattern holds across
all other headline metrics. Macro F1, overall accuracy, $\kappa$ and
frequency-weighted IoU decline for all five models.
All twenty five comparisons therefore show a decline.

\begin{table*}[t]
\caption{Downstream experiment on block fold~1 with seed 42 and 136 real
validation scenes. Both arms use 6000 optimizer steps. R0 contains
804 real pairs, while R1 adds 804 synthetic pairs.
Intervals use 2000 paired bootstrap resamples of validation scenes,
sampled jointly across both arms.}
\label{tbl:stage4}
\begingroup
\setlength{\tabcolsep}{3pt}
\begin{tabular*}{\tblwidth}{@{}L@{\extracolsep{\fill}}CCCCCCCC@{}}
\toprule
\multirow{2}{*}{Architecture} & \multicolumn{4}{c}{Mean IoU} &
\multicolumn{4}{c}{River IoU} \\
\cmidrule(lr){2-5}\cmidrule(lr){6-9}
 & R0 & R1 & $\Delta$ & 95~\% CI & R0 & R1 & $\Delta$ & 95~\% CI \\
\midrule
U-Net (R18)      & 0.4572 & 0.4515 & $-$0.0057 & [$-$0.0130, $+$0.0021] & 0.1587 & 0.1181 & $-$0.0406 & [$-$0.0520, $-$0.0291] \\
UNetFormer (R18) & 0.5055 & 0.4897 & $-$0.0159 & [$-$0.0292, $-$0.0022] & 0.3290 & 0.2281 & $-$0.1008 & [$-$0.1225, $-$0.0778] \\
SegFormer-B0     & 0.4646 & 0.4575 & $-$0.0071 & [$-$0.0180, $+$0.0037] & 0.1463 & 0.1138 & $-$0.0324 & [$-$0.0457, $-$0.0209] \\
SegNeXt-T        & 0.5675 & 0.5311 & $-$0.0364 & [$-$0.0496, $-$0.0214] & 0.4535 & 0.3463 & $-$0.1072 & [$-$0.1310, $-$0.0839] \\
Proposed         & 0.5608 & 0.5560 & $-$0.0047 & [$-$0.0170, $+$0.0078] & 0.4761 & 0.4723 & $-$0.0039 & [$-$0.0416, $+$0.0351] \\
\midrule
Mean of five     & 0.5111 & 0.4972 & $-$0.0140 & --- & 0.3127 & 0.2557 & $-$0.0570 & --- \\
\bottomrule
\end{tabular*}
\endgroup
\end{table*}

\begin{figure*}[!t]
  \centering
  \includegraphics[width=\linewidth]{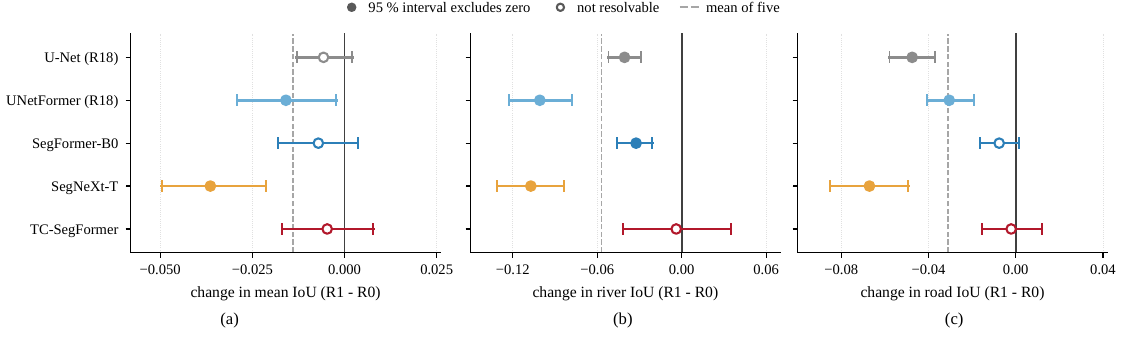}
  \caption{Effect of mixing 804 synthetic image--label pairs with 804
  real pairs over 6000 steps on block fold~1. Changes in
  (a)~mean, (b)~river and (c)~road IoU with 95~\% paired-bootstrap
  intervals. Filled markers indicate intervals excluding zero.}
  \label{fig:downstream}
\end{figure*}

\begin{figure*}[!t]
  \centering
  \includegraphics[width=\linewidth]{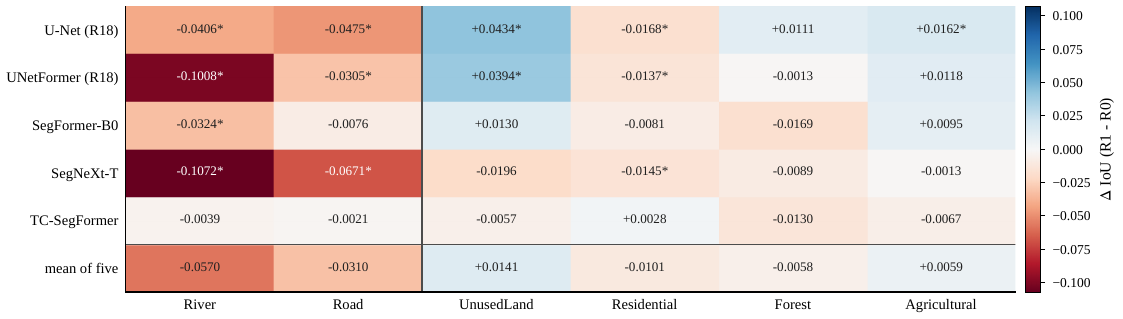}
  \caption{Per-class IoU changes after adding synthetic data across
  architectures. Asterisks mark 95~\% paired-bootstrap intervals
  excluding zero. River and road, left of the rule, are the predefined targets
  of the generative pipeline.}
  \label{fig:perclassdelta}
\end{figure*}

The consistent declines are clear in the interval plot (Fig.~\ref{fig:downstream}).
The per-class results help explain why augmentation failed
(Table~\ref{tbl:stage4class}, Fig.~\ref{fig:perclassdelta}). The pipeline hurts the very
classes it was designed to improve across every architecture. River loses
0.0570 on average across all five models, an 18.2~\% relative decline, with
four of five intervals excluding zero. Road falls by 0.0310, with three
intervals excluding zero. Only unused land and agricultural improve, neither
is a thin structure. Forest shows no meaningful change. Centreline Dice
shows that the losses affect both area and connectivity. It falls for road
in all five models and for river by 15--27~\% relative. All but one
architecture predict river on \emph{more} validation images in R1 than R0,
suggesting that the inflated river presence prior produces false positives.
Frequency-weighted IoU declines by just 40~\% of the
unweighted drop showing that rare classes bear most of the damage rather
than dominant classes.

\begin{table*}[t]
\caption{Per-class IoU changes after adding synthetic pairs and the
  factors linked to them. ``Synthetic'' and ``Real'' give the frozen
  scorer's IoU on generated and real images. Their ratio is synthetic
  divided by real IoU. Contamination is estimated as
  $0.5(1-\text{synthetic IoU})$.}
\label{tbl:stage4class}
\begingroup
\setlength{\tabcolsep}{3pt}
\begin{tabular*}{\tblwidth}{@{}L@{\extracolsep{\fill}}CCCCCC@{}}
\toprule
\multirow{2}{*}{Class} & \multirow{2}{*}{Mean $\Delta$} &
\multirow{2}{*}{CI excl.\ 0} & \multicolumn{3}{c}{Layout fidelity} &
\multirow{2}{*}{Contam.\ (\%)} \\
\cmidrule(lr){4-6}
 & & & Synthetic & Real & Ratio & \\
\midrule
River        & $-$0.0570 & 4$-$ / 0$+$ & 0.0567 & 0.1339 & 0.424 & 47.2 \\
Road         & $-$0.0310 & 3$-$ / 0$+$ & 0.2187 & 0.2991 & 0.731 & 39.1 \\
Residential  & $-$0.0101 & 3$-$ / 0$+$ & 0.8143 & 0.8006 & 1.017 & \phantom{0}9.3 \\
Forest       & $-$0.0058 & 0$-$ / 0$+$ & 0.6757 & 0.7472 & 0.904 & 16.2 \\
Agricultural & $+$0.0059 & 0$-$ / 1$+$ & 0.5436 & 0.5626 & 0.966 & 22.8 \\
Unused land  & $+$0.0141 & 0$-$ / 2$+$ & 0.2950 & 0.2400 & 1.229 & 35.2 \\
\bottomrule
\end{tabular*}
\endgroup
\end{table*}

\begin{figure*}[!t]
  \centering
  \includegraphics[width=364pt]{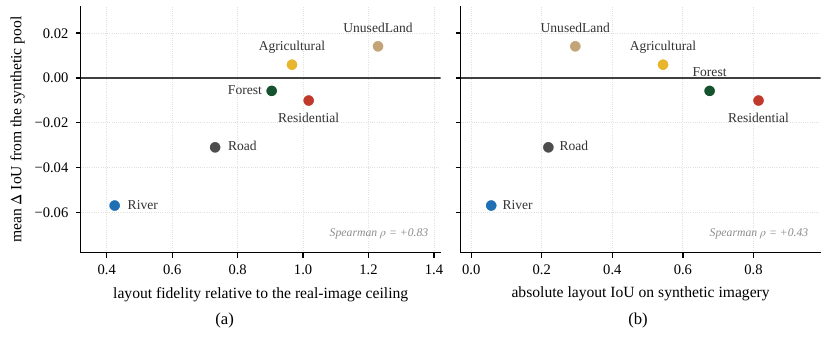}
  \caption{Layout fidelity and downstream performance.
  (a)~mIoU change against layout fidelity relative to the frozen
  scorer's ceiling on real images.
  (b)~mIoU change against absolute layout fidelity on synthetic images.}
  \label{fig:mechanism}
\end{figure*}

The fidelity columns help explain this pattern but the difference between
the measures matters. Ranking the six classes by their \emph{ratio} and
downstream change gives similar results. Using absolute synthetic fidelity
instead shows little relationship. What matters is not how poorly the
generator represents a class but how much worse it performs relative to
the fidelity achievable on real imagery (Fig.~\ref{fig:mechanism}).

Unused land has 35~\% contamination but still improves because it has little
distinctive texture to lose. Its separability from agricultural land is just
0.051 meaning incorrect textures have limited impact, while greater layout variety
provides useful new information. River behaves differently. Its defining
features are a narrow, connected shape and distinct spectral characteristics.
When generated rivers appear as fragmented dark blobs, they introduce
incorrect assumptions about both shape and texture.

Two classes have ratios above 1.0, but this is a warning rather than
an advantage. It does not indicate more realistic generated imagery.
Instead, the frozen scorer identifies these classes more confidently in
generated images than in real ones. The generator produces appearances
closer to the conditional mean, with smoother textures, more typical
features, and cleaner boundaries than real scenes. The ratio therefore
reflects relative recoverability, not realism. Similar saturation occurs
in the condition-source experiment in Section~\ref{sec:results-gen}.
Fig.~\ref{fig:synthetic} shows representative pairs and per-class changes.

\subsection{Efficiency, and the repeatability of throughput}
\label{sec:results-eff}

\begin{figure}[!t]
  \centering
  \includegraphics[width=\linewidth]{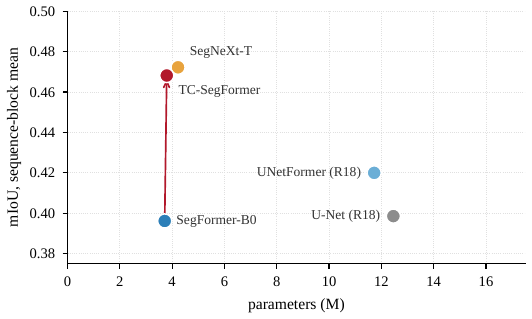}
  \caption{Parameter count versus mIoU across sequence-block folds.
  The arrow shows the proposed model's additional 0.078~M parameters,
  1.9~\% of the SegFormer-B0 backbone.}
  \label{fig:efficiency}
\end{figure}

Each architecture was trained twice under identical conditions, and training
data does not affect inference throughput. Any difference between the two
throughput measurements therefore reflects measurement noise. This variation
ranges from 0.7~\% to 12.4~\%, giving an empirical repeatability limit.
Throughput differences below roughly 12~\% should therefore not be considered
meaningful on this platform. The proposed model uses 3.794~M parameters
compared with 3.716~M for its backbone. Peak training memory also increases
from 6.89~GB to 7.30~GB while throughput remains unchanged
(Fig.~\ref{fig:efficiency}).

\subsection{The cost of variant-level leakage}\label{sec:results-leakage}
\begin{figure}[!t]
  \centering
  \includegraphics[width=\linewidth]{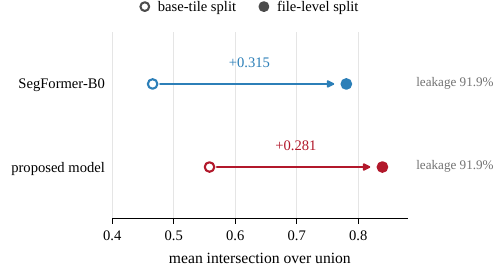}
  \caption{Effect of file-level splitting on mIoU for block fold~1,
  compared with the base-tile split. Arrows show measured variant
  leakage. The two leaked results are not directly comparable.}
  \label{fig:leakage}
\end{figure}

Published land-use segmentation studies using flip augmented datasets often
use file-level splits, allowing mirrored versions of the same scene to appear
in both training and validation sets. Our corpus contains three versions of
each scene (Section~\ref{sec:data-split}) making it possible to measure
the effect of this practice. We retrained SegFormer-B0 and the proposed
model on block fold~1, splitting files instead of base tiles. Measured
variant leakage reached 91.9~\% and 98.8~\%, respectively. These runs also
lasted longer, at 299 and 226 epochs rather than the usual 120, so both
leakage and training duration contribute to the difference. mIoU increased to 0.7806 and 0.7987, compared with the original fold-1
scores of 0.4657 and 0.5582 in Table~\ref{tbl:main}
(Fig.~\ref{fig:leakage}). This represents an increase of roughly 0.25 to
0.30, largely enabled by the split protocol. A model exposed to mirrored
versions of validation scenes can recognise previously seen content.
The higher scores therefore reflect memorisation rather than better
generalisation (Fig.~\ref{fig:qualitative}, rows five and six). The two
leaked runs cannot be compared directly because their leakage levels differ.
We include these results to explain why the scores in
Section~\ref{sec:results-main} are lower than published results using
similar class schemas. The split protocol explains this gap, rather
than differences in the corpus or models. Since each result comes from
a single fold and seed, we report the approximate magnitude without
an interval.

\section{Discussion}\label{sec:discussion}

On this corpus, the data split has a greater impact than the choice of
architecture. Diffusion generated image--label pairs cannot replace real
data. They benefit classes defined by context but harm those relying on
appearance and topology.

\subsection{Three separate causes of rare-class failure}
\label{sec:disc-interp}

\begin{figure*}[!t]
  \centering
  \includegraphics[width=\linewidth]{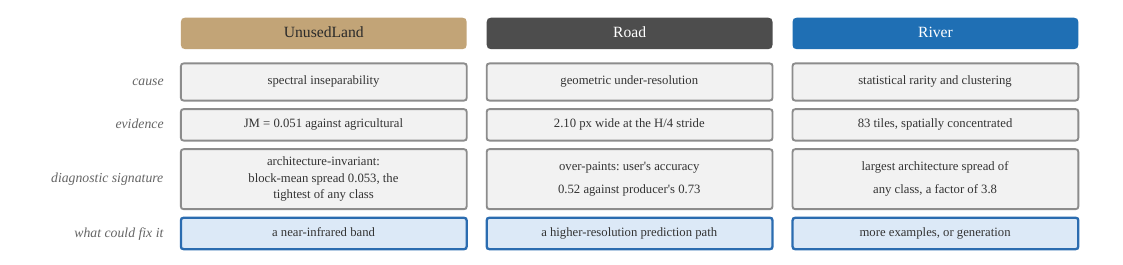}
  \caption{Three separate causes of rare-class failure, with the measurement that
  identifies each and the intervention it implies.}
  \label{fig:failuremodes}
\end{figure*}

River, road and unused land consistently perform worst, suggesting they
share the same problem. However, the results show different causes,
and separating them helps identify which interventions might work
(Fig.~\ref{fig:failuremodes}).

Unused land is difficult because of \emph{spectral inseparability}. Its
Jeffries--Matusita distance from agricultural land is just 0.051. In RGB
imagery, fallow fields and vacant plots look almost identical, while their
distinction depends on context and history rather than pixel values. We
would therefore expect similar performance across architectures, which the
results confirm. The block-mean spread across five different models is
only 0.053, the smallest of any class. The precision/recall gap
(user's accuracy 0.5644 versus producer's 0.3957) suggests that models
make conservative predictions because the classes lack a clear spectral
boundary. Changing the architecture, loss function or synthetic RGB
augmentation cannot overcome this spectral limit.

Road struggles with \emph{geometric under-resolution}. At just 2.10 pixels
wide at the prediction stride, a one-pixel boundary error changes its width
by 50\%. This is a representation problem rather than a learning
problem. It explains why the detail path contributes most in the ablation
with its benefits concentrated on road and river. The precision/recall
results support this: user's accuracy is 0.5229 versus producer's 0.7302.
The model predicts roads too broadly, as expected when narrow structures
are upsampled from a coarse grid.

River struggles with \emph{statistical rarity plus spatial clustering}. Unlike
unused land, it is spectrally distinguishable, and unlike road, it is wide
enough that resolution is not the main problem. Instead, its 83 examples
are correlated and concentrated in a small part of the study area.
This explains why river shows the greatest variation across architectures,
the largest spatial penalty and the most instability between folds.
Among the three classes, river is the only one whose limitations
could plausibly be addressed through generative augmentation.

\subsection{The component ablation}\label{sec:disc-ablation}

Table~\ref{tbl:ablation} shows four leave-one-out runs on block fold~1.
Each component improves performance but the individual gains are small.
Their contributions sum to 0.0284 while the full model exceeds the bare
backbone by 0.0925. The ablation therefore explains less than a third
of the total improvement. This is not a measurement error. It suggests
that the components perform overlapping roles. Removing one has little
effect because the others compensate. Removing all four could have a
much greater impact. The components matter together even though their
individual effects are small. Their contributions should therefore not
be treated as additive. We did not train a model with all four
components disabled. This remains the most useful experiment to
complete in this part of the study.

\begin{table}[t]
\caption{Leave-one-out ablation on block fold~1 with seed 42.
  Deltas are calculated as variant minus full model, with negative
  values indicating a useful component. Per-class changes are shown
  for the three targeted classes.}
\label{tbl:ablation}
\begingroup
\setlength{\tabcolsep}{3pt}
\begin{tabular*}{\tblwidth}{@{}L@{\extracolsep{\fill}}CCCCC@{}}
\toprule
Configuration & mIoU & $\Delta$ mIoU & $\Delta$ River & $\Delta$ Road & $\Delta$ Unused \\
\midrule
Full model        & 0.5582 & --- & --- & --- & --- \\
$-$ detail path   & 0.5449 & $-$0.0133 & $-$0.0392 & $-$0.0295 & $-$0.0180 \\
$-$ sampler       & 0.5509 & $-$0.0073 & $-$0.0030 & $-$0.0075 & $-$0.0262 \\
$-$ soft-clDice   & 0.5522 & $-$0.0060 & $-$0.0162 & $-$0.0159 & $-$0.0004 \\
$-$ reweighting   & 0.5564 & $-$0.0018 & $-$0.0077 & $+$0.0013 & $+$0.0024 \\
\midrule
Sum of the four   & ---    & $-$0.0284 & --- & --- & --- \\
Backbone only     & 0.4657 & $-$0.0925 & --- & --- & --- \\
\bottomrule
\end{tabular*}
\endgroup
\end{table}

\begin{figure*}[!t]
  \centering
  \includegraphics[width=364pt]{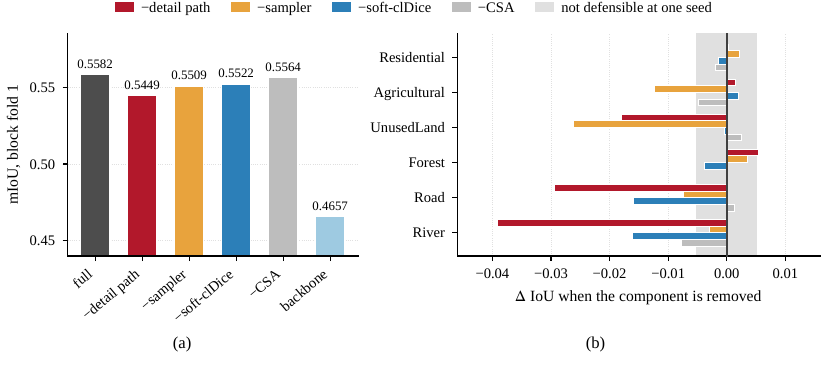}
  \caption{Leave-one-out ablation on block fold~1.
  (a)~mIoU for the full model, each variants with one component disabled,
  and the bare backbone. (b)~Per-class differences.
  The shaded band is the 0.005
  below which nothing is defensible at a single seed.}
  \label{fig:ablationfig}
\end{figure*}

The per-class results tell us more than the overall score
(Fig.~\ref{fig:ablationfig}). Each component mainly helps the classes it was
designed for. The detail path has the largest individual effects on river
and road, while soft centreline Dice improves these two classes with
little effect elsewhere. This is expected since it is applied only
to road and river. The sampler mainly benefits unused land.
Boundary reweighting shows a different pattern. Its overall effect
is $-$0.0018, with a largest per-class effect of $-$0.0077.
Three of the six classes also move in the wrong direction.
With only one run per configuration on a single fold and seed,
there is no clear evidence of improvement. Changes below roughly
0.005 mIoU cannot be reliably distinguished
from noise. This leaves the sampler and topology term near the
threshold while boundary reweighting falls below it.

\subsection{Why synthetic augmentation hurt}\label{sec:disc-synthetic}

Three mechanisms account for the result, and each of them is measured.

The first reason is numerical. In a one-to-one mixture, generated images
provide roughly half the supervision for each class. The absolute layout
fidelity in Table~\ref{tbl:stage4class} indicates how much of that supervision
is usable. Residential has a fidelity of 0.8143 meaning approximately
9~\% of its total supervision is wrong. River has a fidelity of just
0.0567 leaving approximately 47~\% incorrect. The errors are also
systematic rather than random. The painter repeatedly produces wide,
fragmented river shapes instead of narrow, continuous ones. A network
with enough capacity learns these incorrect patterns rather than
averaging them out.

\begin{figure*}[!t]
  \centering
  \includegraphics[width=364pt]{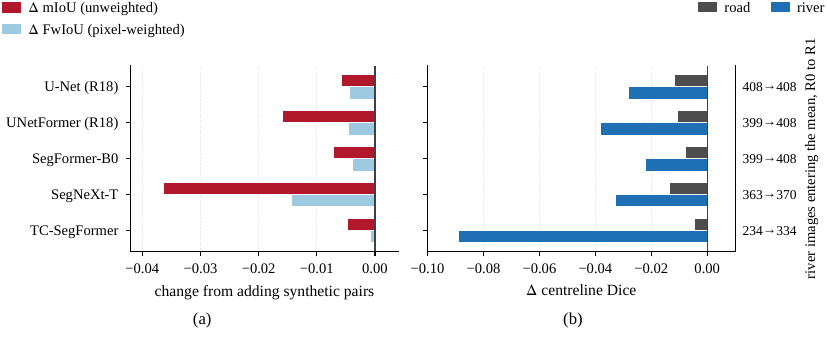}
  \caption{Effect of synthetic data on mIoU and topology.
  (a)~Changes in unweighted beside frequency-weighted IoU.
  (b)~Change in centreline
  Dice for the two thin classes. The right hand axis gives the number of images
  entering each river mean in the two arms.}
  \label{fig:fwiou}
\end{figure*}

The second reason is the presence prior. The pool was matched on class
\emph{pixel} marginals but not class \emph{presence}. The same river pixel
budget is distributed across roughly twice as many scenes, leaving each
river bearing synthetic tile with about 2.7 times less river than a real
one. Models learn this inflated prior and predict river even where none
exists. The attempt to increase rare-class coverage therefore shifts
the model's operating point. Third, our rare-class sampling strategy makes
the problem worse. The presence-balanced sampler calculates frequencies
from real data but samples from a pool containing too many synthetic
river-bearing tiles. As a result, it selects contaminated examples more
often (Fig.~\ref{fig:fwiou}).

Two broader conclusions follow. First, the experiment tests
\emph{substitution at fixed compute} rather than adding synthetic data with
twice the compute. This was intentional because matching epochs would give
the synthetic arm twice as many gradient updates. Our results therefore
show whether synthetic pairs match real pairs per unit of optimisation.
They do not establish whether synthetic data helps when compute is
unlimited. Second, the training curves suggest that several possible
remedies would not work. The train--validation gap \emph{narrowed} in all
five models with no evidence of overfitting. Validation loss is therefore
unreliable for judging rare-class performance here. River performance in
the synthetic arm remained below the real only arm from the first
evaluation and never recovered. The model failed to learn river properly
rather than learning it and later forgetting it. This rules out curriculum
training, late introduction, annealing and early stopping as solutions.
Overall, layout diversity transfers from synthetic data but appearance
fidelity does not. Synthetic pairs therefore benefit classes defined by
context while harming those dependent on appearance and topology.

Our findings differ from \citet{toker2024satsynth}. That study uses a much
larger corpus and achieves appearance fidelity far above 0.42 of the ceiling
for its target classes. Our condition-source ablation shows that the problem
comes from layout generation rather than image synthesis. Our results also
support the concerns about appearance metrics raised by
\citep{kynkaanniemi2023imagenetclasses}. Despite a respectable kernel
Inception distance, our synthetic pool reduces downstream performance.
Realism scores alone would therefore have given a misleading assessment.

\subsection{Limitations}\label{sec:disc-limits}

\begin{itemize}
\item \emph{Single seed.} All results use seed 42 without repeated training.
      The only available noise estimate comes from three evaluations of
      an architecture with stochastic inference. These show evaluation noise
      of at most 0.0006 but provide no estimate of training variance.
      Differences below roughly 0.005 mIoU
      therefore cannot be considered reliable.
\item \emph{The partition is a proxy, not a geometry.}
      Adjacency perpendicular to the scan axis remains uncontrolled. So the
      leakage gap is only a lower bound.
\item \emph{Scale is estimated, not recorded}, within $\pm$1.6~\%.
      This uncertainty affects all metre based statistics.
\item \emph{Annotation quality remains unquantified.} Two annotators drew
      polygon shapes without a second review, and inter annotator agreement
      was not measured. This could still be assessed by independently
      re-annotating a small subset to measure per-class agreement and
      boundary tolerance without changing the released corpus. We consider
      this the most valuable addition to the benchmark. Until then,
      the interior restricted evaluation enabled by our reweighting remains
      diagnostic only. Scoring on eroded interiors would be circular
      without an independently re-annotated reference.
\item \emph{The generative result covers one fold, one seed and one mixture ratio.}
      Downstream training used only the one-to-one mixture without comparing
      it against strong classical augmentation under a matched budget.
\item \emph{RGB only.} The lack of a near-infrared band is the study's
      biggest structural limitation and likely explains the performance
      ceiling for unused land.
\end{itemize}

\section{Conclusion and future work}\label{sec:conclusion}

We released and analysed a six-class land-use corpus covering 202.4~km$^2$
of the Kathmandu Valley. After losing the export geometry, we recovered
its acquisition scale photogrammetrically to $\pm$1.6~\%. We conducted
thirty segmentation fold runs under a common protocol using two partitioning
regimes. The partition matters more than the architecture. Switching from
a random to a sequence-block split reduces mIoU
by 0.051 to 0.095 and increases between fold standard deviation by
a factor of 5.6 to 19.5. We also tested a compact transformer with
a half resolution residual detail path, boundary reliability reweighting,
a topology-aware loss and a presence-balanced sampler. It outperforms
its backbone in all six fold runs by 0.053 to 0.111 with only
1.9~\% more parameters. However, both models perform similarly on LoveDA,
suggesting the improvement is specific to this corpus. Finally, our two stage
diffusion pipeline produces internally consistent image--label pairs without
memorising training samples. Mixing them one-to-one with real pairs at
matched compute reduces mIoU for every tested
architecture by 0.014 on average. The greatest losses occur in the two
rare thin classes the pipeline was designed to improve.

Three steps would help address these limitations. Adding a near-infrared band
for even part of the study area would test the spectral ceiling limiting
unused land. This would be the most valuable addition to the corpus.
Repeating the main comparisons across three seeds, training a configuration
with all components disabled, and comparing synthetic data with strong
classical augmentation at matched compute would provide stronger statistical
evidence for both parts of the study. Our condition-source ablation also
shows that the remaining error comes from layout generation rather than
image synthesis. Future generative augmentation should therefore focus on
layout models that explicitly preserve connectivity and width instead of
improving image synthesis.

For practitioners working with similar data, we recommend three steps.
First, report results under both partitioning regimes because the split
affects performance more than any architectural choice in this study.
Second, select conditional generators using a task relevant measure
rather than denoising loss. Third, before using synthetic image--label
pairs, compare per-class layout fidelity against the real image ceiling.
Classes with ratios far below one are likely to suffer from synthetic
augmentation.

\printcredits

\section*{Declaration of competing interest}

The authors declare that they have no known competing financial interests or
personal relationships that could have appeared to influence the work reported
in this paper.

\section*{Declaration of generative AI use}

During the preparation of this work the authors used Large Language
Models in order to correct English language of some parts of the
manuscript. After using this tool/service, the authors reviewed and
edited the content as needed and take full responsibility for the content
of the published article.

\section*{Code availability}

The training, evaluation and generation notebooks, the dataset-characterisation
and scale-recovery notebooks, and the protocol-deviation ledger referred to in
Section~\ref{sec:method} are available at {\tt https://github.com/Seg-Gen/SegSynth} under the MIT
licence.

\section*{Funding}

This research did not receive any specific grant from funding agencies in the
public, commercial, or not-for-profit sectors.

\section*{Data availability}

The dataset underlying this study is deposited on Zenodo as the Kathmandu Valley
Land-Use and Land-Cover Segmentation Dataset, version~1
\citep{parajuli2026kathmandudataset}, at
{\tt https://doi.org/10.5281/zenodo.22349064}. Access is restricted: the optical
imagery derives from a commercial basemap service whose terms prohibit
redistribution. So, the record is not open and the data are made available on
request for research and study purposes. The record holds the six-class
annotation masks, per-tile metadata and class statistics.

\section*{Acknowledgements}

The authors thank the Department of Electronics and Computer Engineering, IOE Thapathali Campus, for institutional support. They also thank Google for the satellite imagery from which the dataset tiles were derived, and the maintainers of QGIS, CVAT, PyTorch and Hugging Face, on whose freely available tools this work depends.

\bibliographystyle{cas-model2-names}
\bibliography{refs}

\end{document}